\documentclass[10pt,journal,compsoc]{IEEEtran}

\usepackage{framed,multirow}

\usepackage[pdftex]{graphicx}

\usepackage{amssymb}
\usepackage{amsfonts}
\usepackage{amsmath}
\usepackage{latexsym}
\usepackage{footnote}
\usepackage{scalerel}
\usepackage{gensymb}
\usepackage{cite}
\usepackage{ulem}

\usepackage{url}
\usepackage{xcolor}

\begin{document}

\title{CoMo: A novel co-moving 3D camera system}

\author{Andrea Cavagna$^1$, Xiao Feng$^2$, Stefania Melillo$^1$, Leonardo Parisi$^1$, Lorena Postiglione$^1$, Pablo Villegas$^1$
        
    \IEEEcompsocitemizethanks{
        \IEEEcompsocthanksitem $^1$ CNR--ISC (National Research Council - Institute for Complex Systems) UOS Sapienza, Rome, Italy 
    }
    
        \IEEEcompsocitemizethanks{
        \IEEEcompsocthanksitem $^2$ College of Engineering, South China Agricultural University 
    }
    
}

\IEEEtitleabstractindextext{

\begin{abstract} 
Motivated by the theoretical interest in reconstructing long $3D$ trajectories of individual birds in large flocks, we developed CoMo, a co-moving camera system of two synchronized high speed cameras coupled with rotational stages, which allow us to dynamically follow the motion of a target flock. With the rotation of the cameras we overcome the limitations of standard static systems that restrict the duration of the collected data to the short interval of time in which targets are in the cameras common field of view, but at the same time we change in time the external parameters of the system, which have then to be calibrated frame-by-frame. We address the calibration of the external parameters measuring the position of the cameras and their three angles of yaw, pitch and roll in the system \textit{home} configuration (rotational stage at an angle equal to $0\degree$) and combining this static information with the time dependent rotation due to the stages. We evaluate the robustness and accuracy of the system by comparing reconstructed and measured $3D$ distances in what we call $3D$ tests, which show a relative error of the order of  $1\%$. The novelty of the work presented in this paper is not only on the system itself, but also on the approach we use in the tests, which we show to be a very powerful tool in detecting and fixing calibration inaccuracies and that, for this reason, may be relevant for a broad audience.
\end{abstract}

}

\maketitle

\IEEEdisplaynontitleabstractindextext
\IEEEpeerreviewmaketitle

\section{Introduction}

\begin{figure*}[h]
    \centering
    \includegraphics[width=1.0\linewidth]{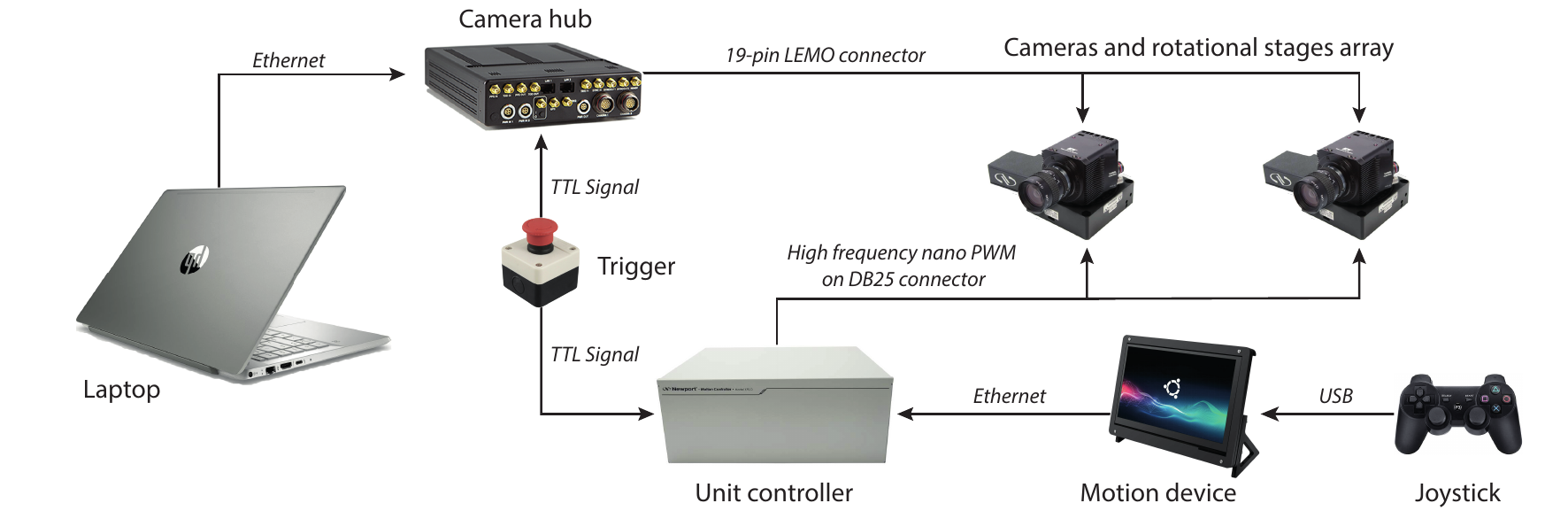}
    \caption{\textbf{Scheme of the system.} The two IDT OS10-4K cameras (resolution $3840px\times 2400px$, sensor size $17.9$mm$\times 11.2$mm, frame rate $155$fps), equipped with Schneider Xenoplan $28$mm f/2.0 optics, are coupled with the two high-speed one-axis rotational stages (Newport RVS80CC, nominal accuracy $10^{-4}$rad, nominal home repeatibility $4\cdot 10^{-3}$rad). Each camera is connected with a 19-pin Lemo cable to the IDT TC-19 hub, which is also connected to a control laptop. The cameras parameters, such as the exposure time, the sensitivity and the frame rate are manually set on the IDT proprietary software Motion Studio running on the laptop. They are sent via an Ethernet connection from the laptop to the hub, which redirects them to the cameras through the 19-pin Lemo cable using an IDT proprietary protocol. The hub sends to the cameras also the synch signal, which is generated by the hub itself. The direction and the speed of rotation of the stages are manually controlled via a Joypad Logitech F310 connected, via a UBS cable, to a motion device, namely a Raspberry Pi 3 Model B+ connected to a 7'' touchscreen. The motion device communicates, via an Ethernet connection, to the unit controller, which redirects the signals to the rotational stages on a DB25 cable, in the form of a high-frequency nano PWNM. The data acquisitions starts with both the cameras and the stages in the \textit{waiting from trigger mode} until they simultaneously receive the trigger signal, a 5V TTL signal. The trigger signal is generated with a standard trigger button, and it is sent at the same time to the hub, which redirects the signal to the cameras, and to the unit controller, which redirects the signal to the stages.
  }
    \label{fig::CoMoScheme}
\end{figure*}
\IEEEPARstart{I}{n} recent years technological advances in the field of imaging and computer vision, together with the growing demand for $3D$ contents, contributed to make digital camera stereo systems accurate in the $3D$ reconstruction and at the same time accessible to a wide audience. This led to the proliferation of stereo vision applications in fields as diverse as entertainment \cite{entertainment2000,entertainment2007, kulshreshth2012evaluating}, surveillance \cite{3Dsurveillance2006, yu2016long, michel2007gpu, wen2017multi}, navigation \cite{navigation2019, murray2000using, broggi2005obstacle, konolige2008outdoor}, robotics \cite{robotics2012, robotics2017, industrial2020, schmid2013stereo}, medicine \cite{medicine2016, fernandes2010stereo, bert2005phantom, probst2017automatic} and biology \cite{attanasi2015GReTA,Theriault2014 ,watts2017validating, cheng2015novel}.
 
The experimental design of a $3D$ system is delicate because of the several factors that contribute to its reliability and feasibility, which strictly depend on the specific data to be gathered and on the environmental and logistic constraints of the data-acquisition location. Standard stereo systems are designed in a static fashion with the position and the orientation of the cameras fixed in time, thus with a fixed field of view. This set-up is suitable for most of the laboratory experiments, where the phenomena to be reconstructed happen in a confined volume, but it represents a severe limitation for non-confined field experiments. 

Ideally when dealing with non-confined phenomena one would like to have a wide field of view and a high resolution of the system. But in fact this is not possible since both factors depend on the cameras focal length, which needs to be short to have a large field of view and it needs to be long to have a high resolution. Therefore one has to lower the data-taking expectations finding a compromise between the two factors, which most of the time ends up reducing both the field of view and the resolution of the system. A smarter, though more complicated, strategy is to replace the static set-up with a dynamic one, effectively widening the field of view with a controlled rotation of the cameras aimed at following the targets, \cite{panningRobotics2006,panningPedestrian2009,panningRobotics2014,panningVehicles2014,panningSurveillance2016,panTilt2014,moving2006}. The dynamic set-up overcomes the limitation of the static one by actually breaking the link between the size of the field of view and the resolution of the system, which is now the only factor depending on the focal length. Hence the resolution of the system can be set as high as needed without reducing the $3D$ volume covered by the system.

The rotation of the cameras makes the external parameters (orientation and position of the cameras in the world reference frame) time-dependent quantities that have then to be carefully calibrated frame by frame to guarantee high accuracy in the reconstruction of the scene. 
The literature suggests two different calibration approaches \cite{CameraCalibration2002}: \textit{i)} $3D$ methods, which reconstruct key points of calibrated $3D$ targets and estimate the external parameters as the ones that minimize the $3D$ reconstruction error \cite{3DmeasureCalibration2000,3DmeasureCalibration2015,3DmeasureCalibration2016,3DmeasureCalibration2019,3DmeasureCalibration2020}; \textit{ii)} $2D$ methods, which match features across the cameras, reconstruct the correspondent $3D$ points that are then projected back on the cameras, and estimate the external parameters as the ones that minimize the reprojection error \cite{2DmeasureCalibration2003, 2DmeasureCalibration2003b, 2DmeasureCalibration2008,2DmeasureCalibration2012,2DmeasureCalibration2015, 2DmeasureCalibration2016, 2DmeasureCalibration2018,2DmeasureCalibration2019a,2DmeasureCalibration2019b}. 
In \cite{2Dvs3D} we show that the two approaches play different roles (both essential and not interchangeable) in the reconstruction process. $2D$ methods give the best performance in matching the images across the cameras, i.e. in the identification of the set of $2D$ points corresponding to the same $3D$ target in different cameras, while $3D$ methods give the best performance in the \textit{actual} $3D$ reconstruction of the identified point-to-point correspondences.

In their standard implementations both the methods start from a set of correspondences, namely $2D$ point-to-point correspondences in the case of $2D$ methods and $2D$ point-to-$3D$ coordinates in the case of $3D$ methods. The reliability of the two methods is then only guaranteed when the starting sets of known $2D$ or $3D$ points cover the entire field of view. This is not problematic for $2D$ methods, where point-to-point correspondences may be found all over the acquired images, but it represents a severe limitation for $3D$ methods in wide field set-up, where it is not always possible to cover the entire field of view with calibrated targets.
A third approach, which is robust with respect to the $3D$ reconstruction accuracy regardless the size of the field of view, consists in calibrating the external parameters of the system by directly measuring the orientation and position of all the cameras in a common reference frame. This latter approach represents a valid alternative to the $3D$ methods described above, but it is generally not used because it requires a particular care in the system set-up that has to be specifically designed to guarantee a precise measurement of the external parameters.

In this paper we present a novel co-moving $3D$ system, CoMo, inspired by the human ability to follow the trajectory of a target with a coordinate movement of the eyes: cameras are coupled with rotational stages that drive a controlled rotation of all the cameras in the same direction and at the same rotational speed, in this way dynamically adapting the field of view to the motion of the targets.

We developed and tested CoMo in the context of $3D$ data-taking of flocks of birds with cameras pointing at a wide region of the sky. This makes $3D$ standard methods for the calibration of the external parameters not appropriate. Therefore, for the calibration of the set of external parameters to be used in the $3D$ reconstruction process, we adopt the direct measure approach, measuring the position and the three angles of yaw, pitch and roll of all the cameras in a common reference frame with the technique described in Section \ref{sec::ExternalCalibration} and Appendix A, while we use the standard $2D$ method described in \cite{attanasi2015GReTA} for the calibration of the parameters used for the identification of point-to-point correspondences across the cameras. We propose also a new procedure to improve the standard calibration of the camera focal length \cite{StandardCalib} which we found to be not sufficiently accurate for our purposes. We discuss this new procedure in Section \ref{sec::CalibrationAccuracy}, where we show how we could detect and fix the inaccuracy on the focal length by performing $3D$ reconstruction tests on calibrated targets.

We extensively tested CoMo to evaluate its performance in terms of the $3D$ reconstruction, see Section \ref{sec::SystemAccuracy} where we show that the comparison between reconstructed and measured $3D$ quantities on calibrated targets gives excellent results with a $3D$ reconstruction error of the order of $1\%$. 

With a full-fledged experimental data-taking campaign in the field we could also check the feasibility of the experiment with CoMo set-up, which proved to be easy to mount and easy to calibrate in the field. The data collected in the field confirmed that with the co-moving strategy we can actually track the flocks significantly longer than with a standard static system, as we show in Video1 of SI.

\section{$\mathbf{3D}$ reconstruction accuracy requirements}\label{sec::requirements}

The requirements on the $3D$ reconstruction accuracy are strictly dependent on the application the data are collected for. We collect field data of bird flocks with the aim of understanding the mechanisms behind the emergence of collective behaviour, and in particular we investigate the correlation properties of these systems \cite{attanasi+al_14, cavagna2015short}, namely we mainly use the data to measure how far (in space) and for how long (in time) the change in the direction of flight of a bird\footnote{The $3D$ velocity vector, $v$, of a bird is given by $\Delta X/\Delta t$ where $\Delta X$ is the $3D$ displacement vector of the bird in the interval of time $\Delta t$. The bird direction of flight, $\hat{v}$, is defined as the velocity versor $\hat{v}=v/|v|$. The change in the direction of the bird is instead given by $\hat{v}-\hat{V}$, where $\hat{V}$ is the direction of flight of the group that is computed averaging the direction of flight of all the birds in the flock. The change in the direction of flight is therefore computed from the distance between the $3D$ position of the bird at time $t$ and $t+\Delta t$.} influences the change in the direction of flight of the other birds in the flock.

In this framework the absolute positions of the birds are not very useful, while the relevant quantities are the birds' directions of flight and the bird-to-bird distances. Therefore we need CoMo to be particularly accurate in the $3D$ reconstruction of the distances between targets. More precisely, we require the relative error on the reconstructed $3D$ target-to-target distances to be: i. not dependent on the position of the targets, to avoid a spatial bias on the quantities we compute; ii. not dependent on the instants of time where the targets live, to avoid a temporal bias on the quantities we compute; iii. smaller than $0.01$, which we define to be the threshold of the accuracy acceptability.

We evaluated CoMo $3D$ reconstruction accuracy with the tests described in detail in Section \ref{sec::test3D}, showing that the system fulfill all the requirements above.

\section{CoMo system}

In this Section we describe the hardware design of CoMo, its field set-up and the calibration procedure we developed to fulfill the $3D$ reconstruction accuracy requirements listed in Section \ref{sec::requirements}. 

\subsection{Design}

The design of CoMo is shown in Fig.\ref{fig::CoMoScheme}: each of the two IDT OS10-4K cameras (resolution $3840px\times 2400px$, sensor size $17.9$mm$\times 11.2$mm, frame rate $155$fps), equipped with Schneider Xenoplan $28$mm f/2.0 optics, is mounted on a high-speed one-axis rotational stage (Newport RVS80CC, nominal accuracy $10^{-4}$rad, nominal home repeatibility $4\cdot 10^{-3}$rad). The cameras are connected to the hub IDT TC-19 that has the double task of redirecting the signals from a laptop controller to the cameras and of synchronizing the cameras via a trigger and a synch signal.

\subsubsection{Motion control}
The rotation of the stages is manually controlled by an operator via a motion device connected to a unit controller (XPS-RL4), which is also connected to the stages. 

The data acquisition procedure starts with cameras and stages in \textit{waiting for trigger} mode, until they simultaneously receive a signal from a hardware trigger connected to the camera hub and to the stage unit controller.
We developed two different motion modes for CoMo:

\textbf{Offline motion mode:} the speed and the direction of rotation are set before the acquisition starts independently for each stage. 

\textbf{Online motion mode:} the speed and the direction of rotation may be chosen online from an operator via a joypad but they are set to be equal for all the three stages (see Appendix B).

The two different motion modes have different applications: we use the offline mode when performing tests on the system, where we need to be versatile on the cameras rotation, while we use the online mode when we collect data in the field and it is of great importance to change the cameras orientations in real time in order to track the moving target.

\subsection{Field set-up}
We perform experiments on bird flocks in the urban environment of Rome, Italy, setting-up CoMo on the roof of Palazzo Massimo alle Terme in front of one of the bigger and more stable birds roosting site in Rome. 

In this location our working distance is of about $150$m with a system baseline, i.e. the distance between the cameras, of about $25$m. The coupling between the cameras (with a sensor size of $17.9$mm$\times 11.2$mm) and the optics (with a focal length of $28$mm) produces a wide field of view of $35.5\degree$ in width and $22.6\degree$ in height. 

\begin{figure}[h]
    \centering
    \includegraphics[width=1.0\linewidth]{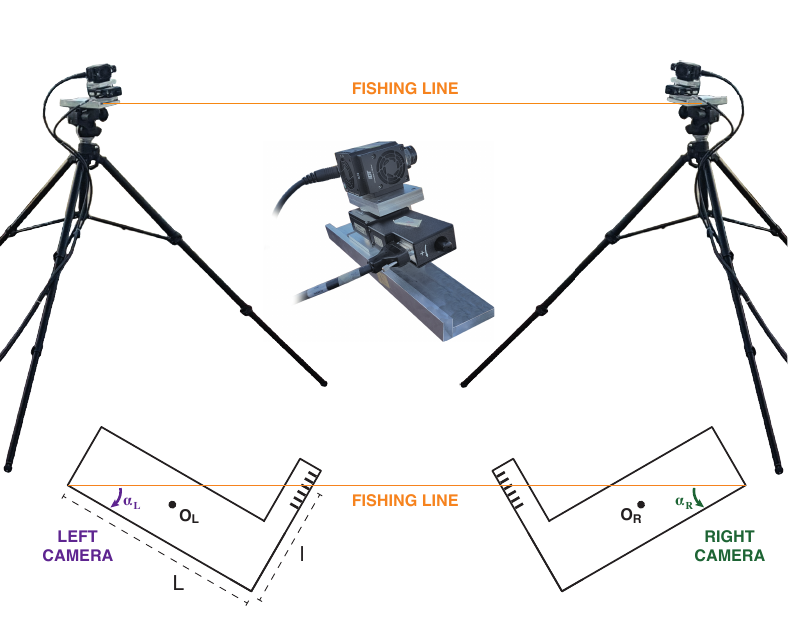}
    \caption{\textbf{Experimental set-up.} Each camera is mounted on a rotational stage that is locked on a L-shape bar and then on a tripod. The L-bars have a gauge on their small edge (on the left side for the right camera and on the right side for the left camera). We set the yaw angles of the cameras by tighten a fishing line, i.e. a thin nylon line, between the two external edges of the bars, so that the fishing line crosses the gauge and it can be used as a pointer on the gauge. Denoting the long side of the L-bar with $L$ and the distance from the point where the line crosses the gauge and the side of the bar with $l$, we can measure the yaw angles as $atan(l/L)$, with the negative sign for the right camera and with the positive sign for the left camera. The accuracy of the measured angle is of $10^{-3}$rad, obtained as $\delta l/L$ with $\delta l$ being the thickness of the wire.}
    \label{fig::experimentalSetup}
\end{figure}{}

\subsection{CoMo calibration}
Our field set-up with the a working distance of $150$m and with a wide field of view of $35.5\degree\times 22.6\degree$ makes the calibration of both the internal and external parameters particularly tough. 

We calibrated the internal parameters, which describe the intrinsic characteristics of the cameras (focal length, position of the image center, distortion coefficients), and external parameters, which define the geometry of the system (orientation and position of all the cameras with respect to a common reference frame in the three dimensional space) with two different procedures.

\subsubsection{Calibration of the internal parameters}\label{sec::InternalCalibration}
For the calibration of the internal parameters we adopt a two-steps procedure. In the first step, we use a standard calibration approach. We calibrate each camera separately in the lab using a standard calibration method based on \cite{StandardCalib}: we collect $50$ images of a $13\times 19$ checkerboard in different positions, we randomly pick $20$ of these pictures and we estimate the focal length, the position of the image center and the first order radial distortion coefficient. We iterate this process $50$ times and we choose each parameter as the median value obtained in the iterations. 

For our dynamic set-up, this standard calibration approach proved to be not accurate enough, producing a time dependent $3D$ reconstruction error due to a slight mis-calibration of the focal length, which we estimated to be of the order of $0.5\%$. Therefore we designed a second step of the calibration, to adjust the focal length using the dynamic approach described in detail in Section \ref{sec::ImprovingOmega}.

Note that, because of the large working distance and of the large field of view, we cannot perform the standard calibration of the internal parameters with a calibration target kept at the working distance, which, at the same time, fills the entire field of view, as this would require a planar target of $98m\times 60m$. Therefore, we chose to reduce the distance of the calibration target in favor of filling the field of view. 

This might be the reason for the mis-calibration of the cameras focal length obtained with the standard method in the first step of our calibration procedure. However, our results are also compatible with a different scenario, which may be the scope of interesting future investigation: the standard calibration approach is less sensitive than the dynamic one to small variations of the estimated focal length; hence, these variations are boosted, and therefore more detectable, using dynamic information. This latter scenario suggests that the dynamic approach to the calibration may be an efficient and relatively simple strategy to improve the calibration performance both for static and for dynamic system configuration.

\subsubsection{Calibration of the external parameters}\label{sec::ExternalCalibration}
In \cite{2Dvs3D} we point out the need of two different sets of external parameters: a first set to be used to match points across the cameras, and a second set to be used in the $3D$ reconstruction process. For the calibration of the first set of parameters we use a standard $2D$ calibration procedure, and we refer the interested reader to \cite{attanasi2015GReTA}, while here we focus on the calibration of the second set of parameters, i.e. the one used for the $3D$ reconstruction.

Our experimental set-up, with a working distance of $150$m and with a field of view, is not suitable to calibrate the external parameters with standard procedures. Our field of view is essentially a wide area of the sky, where we cannot locate any calibration $3D$ target, hence we cannot use a $3D$ calibration method. We prefer to not use feature-based calibration routine because of their low accuracy in the $3D$ reconstruction, which we show to be higher than $1\%$ in \cite{2Dvs3D}. Therefore we address the calibration with a different strategy. 

CoMo external parameters are actually given by the combination of a static term, which does not change in time and describes the initial position/orientation of the cameras, and of a dynamic term, which is time-dependent and describes the rotation of the cameras due to the stages. We directly measure these two terms separately. 

We initially set the rotational stages in their \textit{home} position, i.e. angle of rotation equal to $0$rad, and we set the pitch and roll angles of both cameras respectively to $0$rad and $0.22$rad using a clinometer (RS Pro Digital level 667-3916, accuracy $3\cdot 10^{-3}$rad). We set the yaw angle of the left camera, $\alpha_L$, to $0.11$rad and the yaw angle of the right camera, $\alpha_R$, to $-0.11$rad with a simple but effective technique, see Fig.\ref{fig::experimentalSetup} and Appendix A, with which we achieve an accuracy of $10^{-3}$rad and which we extensively tested on static camera systems, \cite{cavagna2008} and \cite{cavagna2015error}. With this procedure we measure the orientation of both cameras in a common reference frame but, to define the positions of the two cameras in the \textit{real} $3D$ world, we still need to fix a metric scale factor that we calibrate by measuring the system baseline, i.e. the distance between the cameras, with a high precision range finder (Hilti Laser PD-E, accuracy $1$mm).

We start the data acquisition moving the cameras from this \textit{home} calibrated configuration, recording the time-dependent angles of rotation of the stages. With a post-processing procedure we can then associate to each camera frame the correspondent external parameters combining the external parameters measured in the \textit{home} configuration and the time-dependent rotation of the stages recorded during the data acquisition, as described in Section \ref{sec::externalParameters} and Section \ref{sec::timeDiscretization}.

\section{Dynamic $\mathbf{3D}$ reconstruction}
There is a vast literature about $3D$ reconstruction for static camera systems, i.e. system with fixed cameras orientation, \cite{3DmeasureCalibration2015,3DmeasureCalibration2016,3DmeasureCalibration2019,3DmeasureCalibration2020,2DmeasureCalibration2003b,2DmeasureCalibration2008,2DmeasureCalibration2015,2DmeasureCalibration2016,2DmeasureCalibration2019a,2DmeasureCalibration2019b}. Here we move a step forward to generalize the $3D$ reconstruction theory to our dynamic system.

\subsection{Camera reference frame} 
\begin{figure}[h]
    \centering
    \includegraphics[width=0.7\linewidth]{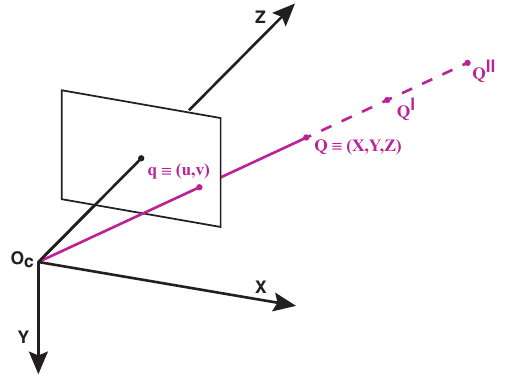}
    \caption{\textbf{Single camera.} The camera reference frame has the origin in the camera optical point, $O_{\scaleto{C}{4pt}}$ the $z$-axis directed as the optical axis and the $xy$-plane parallel to the sensor with the $x$-axis pointing right and the $y$-axis pointing down. The pinhole model describes the relation between the $3D$ world and the $2D$ camera sensor as a central projection: the image $q$ of a $3D$ point lies at the intersection between the sensor and the line passing through $Q$ and the camera optical center. This correspondence is not one-to-one, because $q$ is not only the image of the point $Q$, but also of all the other $3D$ points belonging to the optical line, $O_{\scaleto{C}{4pt}}Q$. This ambiguity makes a single camera not sufficient for the $3D$ reconstruction.}
    \label{fig::SingleCamera}
\end{figure}

The camera reference frame $O_{\scaleto{C}{4pt}}xyz$ has the origin, $O_{\scaleto{C}{4pt}}$, in the camera optical point, the $z$-axis directed as the optical axis and the $xy$-plane parallel to the sensor with the $x$-axis pointing right and the $y$-axis pointing down, see Fig.\ref{fig::SingleCamera}. In our dynamic set-up this reference frame is not fixed in time but it rotates on the $xz$-plane around the camera optical center.

\subsection{Pinhole model}
The pinhole camera model describes the mapping between the $3D$ real world and the $2D$ camera world as a central projection, see Fig.\ref{fig::SingleCamera}: the $2D$ image, $q$, of the $3D$ point $Q$ lies at the intersection between the camera sensor and the line between $Q$ and the camera optical center, $O_{\scaleto{C}{4pt}}$. Its natural mathematical framework is then projective geometry, where the correspondence between a $3D$ point $Q\equiv(X,Y,Z)$ and its $2D$ image $q\equiv(u,v)$\footnote{For the sake of simplicity in the manuscript we will refer to the $2D$ coordinate of an image point as defined in the image reference frame with the origin in the image center instead of the standard reference with the origin in the top left.} is expressed in a very simple formalism:
\begin{equation}\label{eq::P_singleCamera}
    \mathbf{q} = P\cdot\mathbf{Q}
\end{equation}

where: $\mathbf{q}=(\bar u,\bar v,\bar w)$ is the $2D$ projective point corresponding to $q$, namely $u=\bar u/\bar w$ and $v=\bar v/\bar w$, $\mathbf{Q}=(X,Y,Z,1)$ represents the homogeneous projective coordinates of $Q$, \cite{hartley2003multiple}. $P$ is the $3\times 4$ matrix of the form $P=K\cdot[R|T]$, where $K$ is the $3\times 3$ matrix of the camera internal parameters, $R$ and $T$ are respectively the $3\times 3$ rotation matrix and the three components translation vector that bring the camera reference frame in the world reference frame where $Q$ lives, and they both depend on the external parameters of the system. 

This definition of $P$ can be further simplified by noting that 
\begin{equation}\label{eq::T}
T=-R\cdot C
\end{equation}
where $C$ is the vector that connects the origin of the world reference frame to the origin of the camera reference frame, hence $P=KR\cdot[I|-C]$ where $I$ denotes the $3\times 3$ identity matrix. 

In a static camera both $R$ and $C$ are fixed in time, but in our dynamic system the camera reference frame rotates about the camera optical center. Hence the vector $C$ is constant in time, while $R\equiv R(t)$. The time-dependent generalization of the projective matrix is then straightforward:
\begin{equation}\label{eq::P(t)}
P(t)=KR(t)\cdot[I|-C]
\end{equation}

\subsection{World reference frame}\label{sec::worldReferenceFrame}
\begin{figure}[h]
    \centering
    \includegraphics[width=1.0\linewidth]{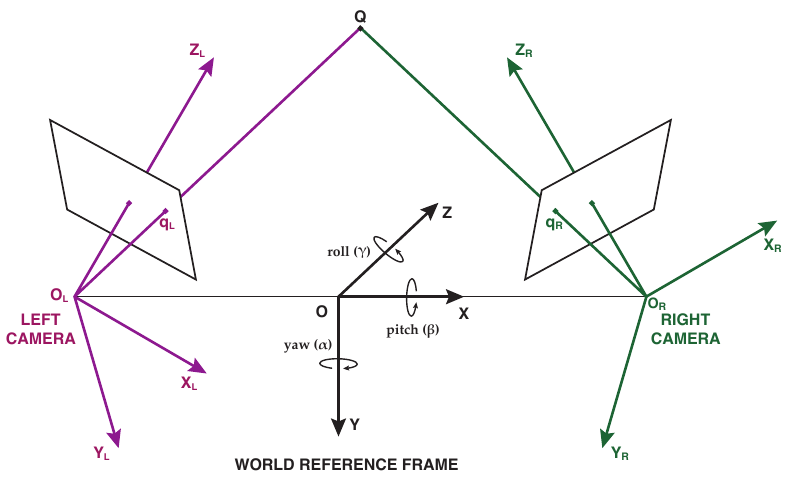}
    \caption{\textbf{Camera system.} $O_{\scaleto{L}{4pt}}x_{\scaleto{L}{3pt}}y_{\scaleto{L}{3pt}}z_{\scaleto{L}{3pt}}$ and $O_{\scaleto{R}{4pt}}x_{\scaleto{R}{3pt}}y_{\scaleto{R}{3pt}}z_{\scaleto{R}{3pt}}$ represent the left and the right camera reference frames. $Oxyz$ is instead the world reference frame, with the origin on the middle point of the camera baseline, $O_{\scaleto{L}{4pt}}O_{\scaleto{R}{4pt}}$, the $x$-axis pointing towards $O_{\scaleto{R}{4pt}}$, the $y$-axis pointing down along the world gravity axis and the $z$-axis pointing outward following the right hand rule. In this reference frame the coordinates of the two camera centers are $C_{\scaleto{L}{4pt}}=(-d/2,0,0)$ and $C_{\scaleto{R}{4pt}}=(d/2,0,0)$. The circle arrows specify the positive direction and the axis of rotation for the yaw, pitch and roll angles.}
    \label{fig::CameraSystem}
\end{figure}

We denote the 2 cameras reference frames by $O_{\scaleto{L}{4pt}}x_{\scaleto{L}{3pt}}y_{\scaleto{L}{3pt}}z_{\scaleto{L}{3pt}}$ (for the left camera) and $O_{\scaleto{R}{4pt}}x_{\scaleto{R}{3pt}}y_{\scaleto{R}{3pt}}z_{\scaleto{R}{3pt}}$ (for the right camera). We define also a third reference frame, $Oxyz$, with the origin on the middle point of the camera baseline, $O_{\scaleto{L}{4pt}}O_{\scaleto{R}{4pt}}$, see Fig.\ref{fig::CameraSystem}, the $x$-axis pointing towards $O_{\scaleto{R}{4pt}}$, the $y$-axis pointing down along the world gravity axis and the $z$-axis pointing outward following the right hand rule. This reference frame is fixed in time, and this is the reference frame within which we will reconstruct the scene. It is then in this reference frame that we need to express the projective matrices of the two cameras.

\subsection{External parameters}\label{sec::externalParameters}
As we already stated in Section \ref{sec::ExternalCalibration}, with our set-up the external parameters are the combination of a static term, that describes the \textit{home} configuration, and of a dynamic term, that describes the rotation due to the stages,
thus the camera rotational matrices are of the form:
\begin{equation}\label{eq::R}
    R_{\scaleto{C}{4pt}} = R_{y_{\scaleto{C}{3pt}}}(-\varphi_{\scaleto{C}{4pt}}(t))\cdot R_S(\alpha_{\scaleto{C}{4pt}},\beta_{\scaleto{C}{4pt}},\gamma_{\scaleto{C}{4pt}})
\end{equation}
where the subscript $C$ indicates a generic camera (left or right), $R_{y_{\scaleto{C}{3pt}}}(-\varphi_{\scaleto{C}{4pt}}(t))$ is the time-dependent rotation, about the $y$-axis of the camera reference frame, which takes into account the rotation of the stage of an angle $\varphi_{\scaleto{C}{4pt}}(t)$, $R_S(\alpha_{\scaleto{C}{4pt}},\beta_{\scaleto{C}{4pt}},\gamma_{\scaleto{C}{4pt}})$ is the static rotation matrix that takes into account the \textit{home} orientation of the camera and $\alpha_{\scaleto{C}{4pt}}$, $\beta_{\scaleto{C}{4pt}}$ and $\gamma_{\scaleto{C}{4pt}}$ are the angle of yaw, pitch and roll respectively\footnote{In the world reference frame the $x$-axis is parallel to the fishing line that we use to measure the two yaw angles, $\alpha_{\scaleto{L}{4pt}}$ and $\alpha_{\scaleto{R}{4pt}}$ (see Fig.\ref{fig::experimentalSetup}), which are then automatically measured with respect to the world reference frame. A similar argument holds also for the pitch and roll angles that we measure using a clinometer because the $y$-axis of the world reference frame is parallel to the gravity direction.}. In particular, for our system 
\begin{equation}\label{eq::RS}
R_S=R_{z_{\scaleto{C}{3pt}}}(-\gamma_{\scaleto{C}{4pt}})\cdot R_{x_{\scaleto{C}{3pt}}}(-\beta_{\scaleto{C}{4pt}})\cdot R_{y_{\scaleto{C}{3pt}}}(-\alpha_{\scaleto{C}{4pt}})
\end{equation}
Note that the order of the rotations in eq.(\ref{eq::RS}) is crucial and it explicitly depends on the tripod model used in the experimental setup, see Appendix A.

Note also that our choice of the world reference frame, with the origin at the center of the camera baseline and the $x$-axis pointing towards the right camera, makes the expression for the two camera centers, $C_{\scaleto{L}{4pt}}$ and $C_{\scaleto{R}{4pt}}$, extremely convenient: $C_{\scaleto{L}{4pt}} = (-d/2,0,0)$ and $C_{\scaleto{R}{4pt}} = (d/2,0,0)$ with $d$ being the length of the baseline.

\subsection{Time discretization}\label{sec::timeDiscretization}
In the previous section we derived the expression of the cameras rotational matrices implicitly considering the time as a continuous variable, while in the actual experimental set-up time is in fact measured in discrete steps, what we normally call \textit{frames}.

\begin{figure}[h]
    \centering
    \includegraphics[width=1.0\linewidth]{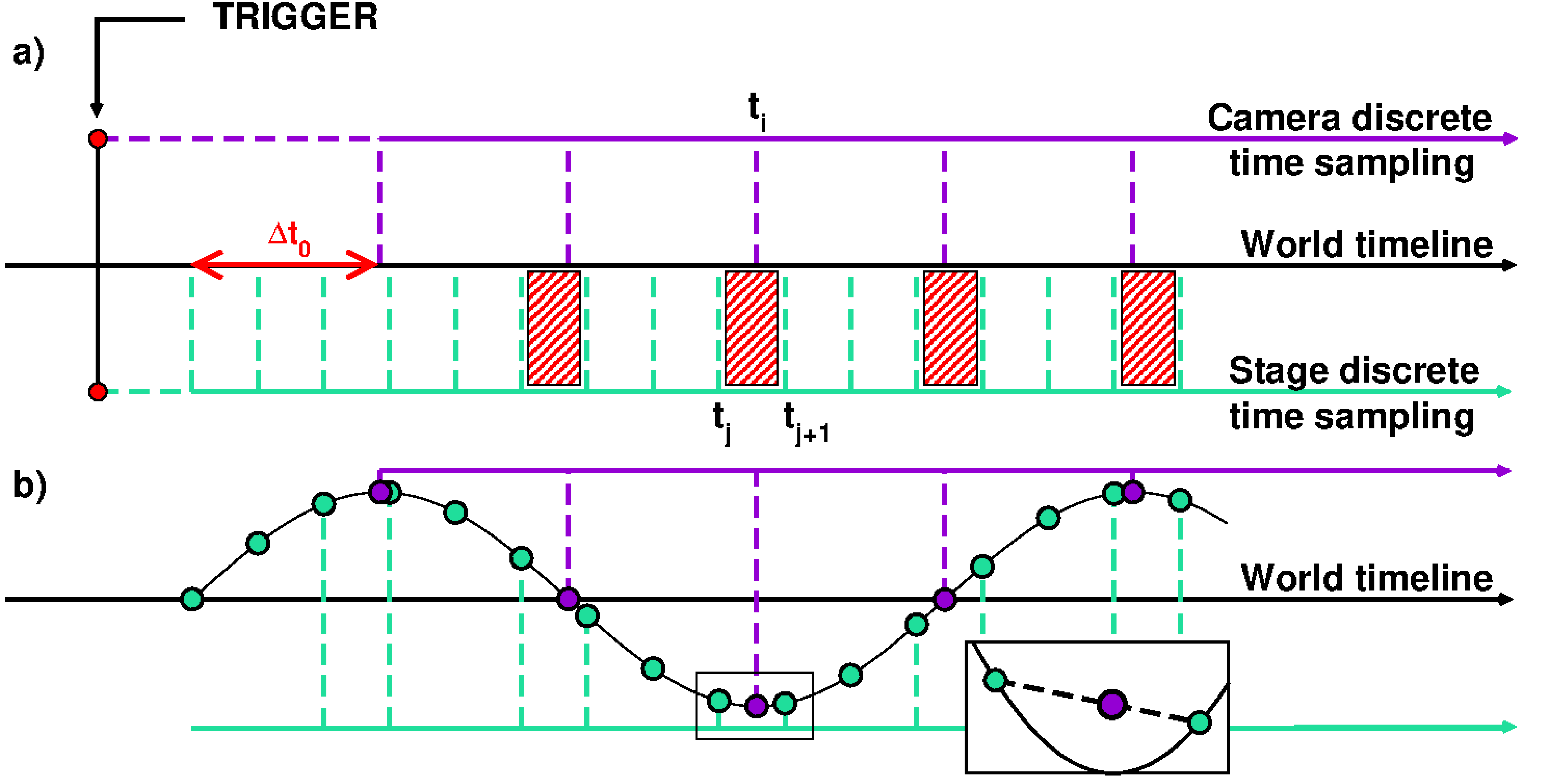}
    \caption{\textbf{Time discretization.} \textbf{a.} The two time discretization of the camera (purple dashed line) and of the stages (green dashed line) have to be matched to associate the position of the stage at each camera sample, i.e. frame. In the continuous world timeline, cameras and stages receive the trigger signal simultaneously, but due to hardware lag time they do not start to record immediately and in general not at the same time. We do not need to know the recording starting time of cameras and stages in the world timeline, but we need to measure the camera-stage offset (red double arrow on the world timeline axis). Once we know this offset we can match each camera time sample, $t_i$, with its two closest time samples of the stage, $t_j$ and $t_{j+1}$: $t_i\in[t_j, t_{j+1}]$. This intervals are highlighted with white and red striped boxes. \textbf{b.} The black sinusoidal line represent the angle of rotation of the stage. The green circles correspond to the time samples of the stage, where the angle is actually measured, while the purple circles correspond to the camera time samples where we need to know the stage position. We associate to each camera time sample the angle obtained with a linear interpolation at time $t_i$ between the two points $(t_j,\varphi(t_j))$ and $(t_{j+1}, \varphi(t_{j+1}))$.}
    \label{fig::TimeDiscretization}
\end{figure}

In a standard static system the only relevant time rate is the one of the cameras. Time discretization can then be efficiently addressed by expressing all the dynamic quantities in the camera frame unit of time. In our dynamic system we have instead two time rates: the one of the cameras defined by the cameras frame rate and the one of the rotational stages defined by their sampling rate. Cameras and stages discretize time with two different rates (the cameras shoot at $155$fps and the stages gather the data at $1000$Hz), see Fig.\ref{fig::TimeDiscretization} where the camera and the stage sampling times are highlighted with purple and light green dashed lines respectively. We reconstruct the position of the targets from the images, hence our primary time rate is the one of the cameras. In order to perform an accurate calibration of the external parameters we need to match this primary time line with the secondary time line of the stages and associate the correct stage position at each camera frame. 

In addition to these two discretizations of time, we have also the continuous world time line. In the world time line, cameras and stages receive the trigger signal simultaneously but due to hardware time lags, which are different for the cameras and for the stages, they do not start to record immediately and in general not at the same time. We do not need to know the recording starting times with respect to the world reference, but it is crucial to know the time delay between cameras and stages, $\Delta t_0$, highlighted with a red arrow on the world timeline in Fig.\ref{fig::TimeDiscretization}. We measured $\Delta t_0$ with the procedure described in Section \ref{sec::timeOffset} and we estimated a delay of $3$ms of the cameras with respect to the stages.

Once this time offset is measured we can express the time corresponding to the camera frame and the time corresponding to the stage samples in the same reference, defining the $i$-th camera time as $t_i=\Delta t_0 + i\Delta t_{\scaleto{C}{4pt}}$ and the $j$-th stage time as $t_j=j\Delta t_{\scaleto{S}{4pt}}$, where $\Delta t_{\scaleto{C}{4pt}}=1/155$s and $\Delta t_{\scaleto{S}{4pt}}=1/1000$s denote the time step of the cameras and the stages respectively.

Finally we associate to the $i$-th camera frame, $t_i$, its two closest stage samples, $t_j$ and $t_{j+1}$, such that $t_i \in [t_j, t_{j+1}]$ see Fig.\ref{fig::TimeDiscretization} where these last intervals are highlighted with white and red striped boxes, and we define the angle $\varphi_C(t_i)$ with a linear interpolation of the two angles $\varphi_C(t_j)$ and $\varphi_C(t_{j+1})$ measured by the stages.

\subsection{$\mathbf{3D}$ reconstruction} 
The ambiguity of the camera projection, which associates to the same $2D$ image all the $3D$ points lying on the same optical line shown in Fig.\ref{fig::SingleCamera}, can be solved with two cameras, see Fig.\ref{fig::CameraSystem}: if $q_{\scaleto{L}{4pt}}$ and $q_{\scaleto{R}{4pt}}$ are the images of the same point, $Q$, in the left and the right camera, $Q$ must lay on the two optical lines, one for each camera, passing through the two images and it is then the point at the intercept between the two lines. In a mathematical formalism this consists in solving the following system in the unknown $\mathbf{Q}$:
\begin{equation}\label{eq::3D}
    \left\{
    \begin{array}{cc}
         \mathbf{q_{\scaleto{L}{4pt}}} = & P_{\scaleto{L}{4pt}}(t)\cdot\mathbf{Q} \\
         \mathbf{q_{\scaleto{R}{4pt}}} = & P_{\scaleto{R}{4pt}}(t)\cdot\mathbf{Q}  
    \end{array}
    \right.
\end{equation}
where $\mathbf{q_{\scaleto{L}{4pt}}}$ and $\mathbf{q_{\scaleto{R}{4pt}}}$ are the $2D$ projective points corresponding to $q_{\scaleto{L}{4pt}}$ and $q_{\scaleto{R}{4pt}}$ and $\mathbf{Q}=(X, Y, Z, 1)$ is the $3D$ homogeneous projective point corresponding to $Q$. $P_{\scaleto{L}{4pt}}(t)$ and $P_{\scaleto{R}{4pt}}(t)$ are the projective matrices of the left and the right cameras defined as in eq.(\ref{eq::P_singleCamera}), each with its own calibration matrix, $K_{\scaleto{L}{4pt}}$ and $K_{\scaleto{R}{4pt}}$, its own rotation matrix defined as in eq.(\ref{eq::R}), $R_{\scaleto{L}{4pt}}$ and $R_{\scaleto{R}{4pt}}$, and its own center in the world reference frame, $C_{\scaleto{L}{4pt}}$ and $C_{\scaleto{R}{4pt}}$.

In deriving eq.(\ref{eq::3D}) we assumed that at each instant of time we detect the exact position of the targets on the images, without considering any kind of noise. The direct effect of noise is that the two lines defined by system (\ref{eq::3D}) do not intersect anymore. Therefore the $3D$ reconstructed coordinates cannot be found as the exact solution of the system but as its approximation, which we obtain using the standard DLT (direct linear triangulation) method in \cite{hartley2003multiple}.

Note that in eq.(\ref{eq::3D}) we identify the camera positions with the optical centers, even though we do not know their exact position. We assume that the optical centers are located in the same position on the camera body (except for small fluctuations), because the factory design is the same for both cameras. We assume also that they are both located at the center of the camera body, which may be not completely correct. With this choice we may then produce a mis-position of the two cameras, which in principle may affect the $3D$ reconstruction accuracy of the system. But the error that we are introducing is a systematic error, i.e. equal for both cameras, hence we may induce a systematic error on all the $3D$ reconstructed points, namely a \textit{solid translation} of the $3D$ world. This may be relevant for the accuracy on the absolute position of the targets, but it does not affect the accuracy on the mutual distance between pairs of targets, which is what we are interested in, as we stated in Section \ref{sec::requirements}.


\section{Time discretization: tests.}

We extensively tested the equipment to measure the time-offset $\Delta t_0$, defined in Section \ref{sec::timeDiscretization} and highlighted with a red arrow in Fig.\ref{fig::TimeDiscretization}, and to check the consistency of the cameras frame rate and of the synchronization between the cameras.

\subsection{Time offset}\label{sec::timeOffset}
\begin{figure}[!h]
\centering
\includegraphics[width=1.0\linewidth]{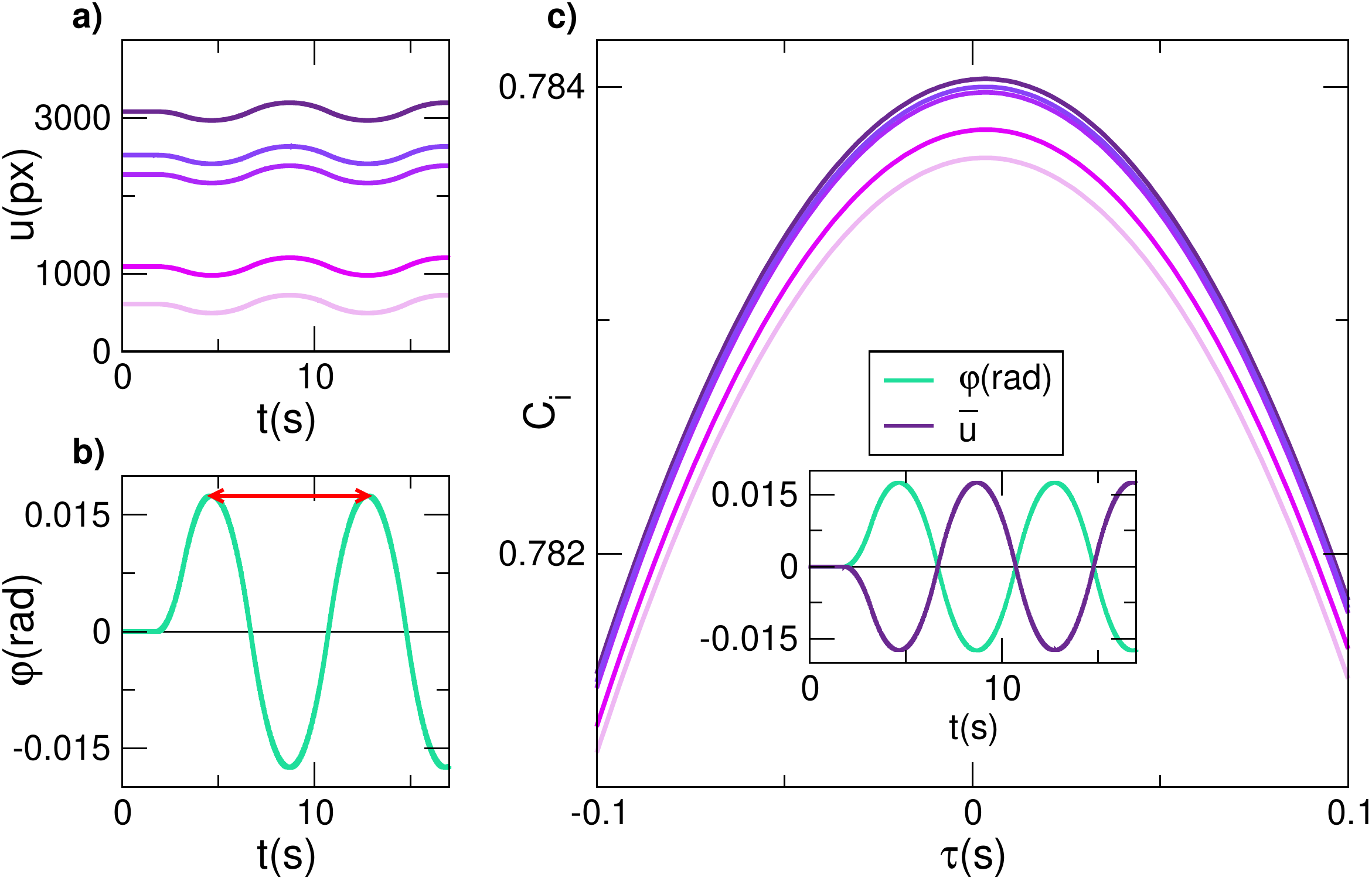}
\caption{\textbf{Time offset.} In order to measure the camera-stage time offset, we acquired images of 5 different targets while rotating the cameras with a periodic movement between $1\degree$ and $-1\degree$. The targets are still, hence the rotation of the cameras due to the stages produces an apparent rotation of the 2D coordinates of the targets at the same speed but in the opposite direction. We estimate the offset from the cross-correlation between the signal recorded by the stage and the position of the targets. \textbf{a.} The evolution in time of the position of the five targets used in the test, each highlighted in a different color. \textbf{b.} The signal recorded by the rotational stage. \textbf{c.} The correlation function $C_i(\tau)$ for each of the five  targets. The maximum of all the cross-correlation functions occur at the same time, which is the offset $\Delta t_0$. In the inset the angle recorder by the stage, green line, and the position of one of the targets, purple line, normalized to be represented on the same scale in the plot. The comparison between the two signals shows the apparent movement of the target at the same speed of the stage but in the opposite direction.}
\label{fig::offset}
\end{figure}
We measured the time offset between the cameras and the stages recording images of five targets ($2\times2$ cardboard checkerboards) while rotating the stages with a periodic movement between $1\degree$ and $-1\degree$, starting with the stages in their \textit{home} position. 

The targets are still, hence the rotation of the cameras (due to the stages) produces an apparent rotation of the 2D coordinates of the targets: if a camera rotates in the clockwise direction at a certain speed, we will detect a rotation of the $u$-coordinate of the targets with the same speed but in the counterclockwise direction and vice versa a counterclockwise rotation of the camera corresponds to a clockwise rotation of the targets. Therefore we can estimate the time offset comparing the signal gathered by the stages with the position of the targets, see Fig.\ref{fig::offset}a where we plot the $u$-coordinates of the five targets, and Fig.\ref{fig::offset}b where we plot the angle recorded by the stage as a function of time. To this aim we compute the cross-correlation of the two signals, taking care of the following three factors: i. the two signals are recorded with different time discretization; ii. the duration of the signals is finite in time; iii. targets positions are not centered in $0$.

We over-sampled the signal from the cameras, i.e. the $u$-coordinate of the targets, with a linear interpolation. In this way we resampled the camera signal at $1000$Hz (the gathering frequency of the stages), so that the time resolution of the cross correlation is defined by the time discretization of the rotational stage. We took care of the finite duration of the two signals restricting the signal of the stage at one period (from the first to the second maximum), highlighted with a red arrow in Fig.\ref{fig::offset}b. Finally we normalized the target coordinates by subtracting their \textit{home} position, see the inset of Fig.\ref{fig::offset}c where we plot the signal from the stage in light green and in purple the position of one of the targets, normalized to be on the same $y$-scale of the stage.

We define the cross-correlation between the signal of the stage and the coordinate of the $i$-th target as:
\begin{equation}
    C_i(\tau) = \displaystyle\frac{1}{T-\tau}\sum_{t=0}^T\varphi(t)\bar{u}_i(t+\tau)
\end{equation}
where $\bar{u}_i$ is the normalized position of the $i$-th target. For each target we can define $\tau_i$ as the point where $C_i(\tau)$ reaches its maximum. We found that all the targets have the maximum of $C_i(\tau)$ at the same point, see Fig.\ref{fig::offset}c, which is the time offset $\Delta t_0$ between the cameras and the stages and that we estimated to be equal to $3$ms.

\subsection{Frame rate consistency and cameras synchronization}
We checked the frame rate consistency and the synchronization between the cameras using a chronometer that we built specifically for these tests: a needle spins at a constant rotational velocity ($20$rps) over a protractor. Knowing the rotational speed of the needle, frame rate and synchronization accuracy are then directly measured from the angle between the position of the needle in two different images. 

We tested the frame rate consistency for each camera separately, by measuring the angle span by the needle between two subsequent images. We found a negligible error, i.e. the error is below our resolution of $7\cdot10^-5$s corresponding to $0.5\degree$ at a rotational speed of $20$rps. We also tested the synchronization between the cameras, comparing the position of the needle on the images acquired at the same time frame from different cameras and again we found a negligible error. 

\section{Yaw angles accuracy in time}

The accuracy of the time dependent yaw angles, $\varphi_L(t)$ and $\varphi_R(t)$, depends on two factors, namely the rotational stage home repeatibility and the accuracy on the interpolation we use to compute $\varphi_L(t)$ and $\varphi_R(t)$, described in Section \ref{sec::timeDiscretization}. We evaluated the accuracy both on the home repeatibility and on the interpolation on each pair camera/stage separately, by preforming the tests shown in this Section.

\subsection{Stage home repeatability}
\begin{figure}[!h]
\centering
\includegraphics[width=1.0\linewidth]{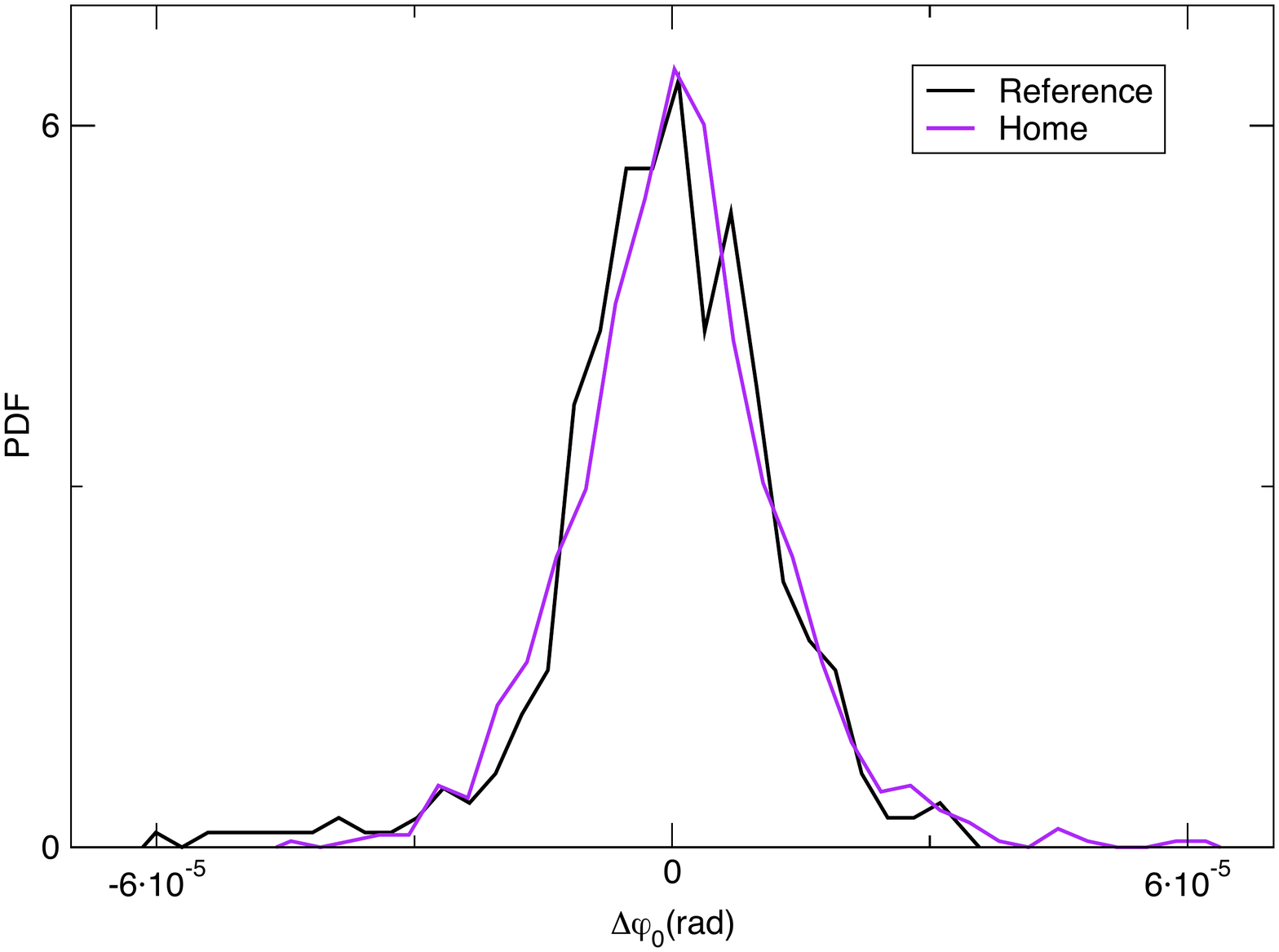}
\caption{\textbf{Home repeatibility.} The probability distribution function (PDF) obtained while acquiring images with the stage still represents the reference distribution for our test. The reference distribution, highlighted in black, gives the measure of the fluctuations due to the targets detection routine. The PDF of the fluctuations on the home position of the stage, highlighted in purple, obtained homing the stage after its initialization procedure. The PDF is compatible with the reference distribution, with the fluctuations smaller than $6\cdot 10^{-5}$rad and with a zero median value.}
\label{fig::home_repeatibility}
\end{figure}

In the rotational stage \textit{home} procedure we include the initialization of the stage, namely we first initialize the stage and then we move it to the home position. The unit controller offers also a direct procedure to home the stage from a generic position, but we chose the indirect procedure because of its higher consistency, see Appendix C for more details.

We denote by $\varphi_0$ the stage home position. We cannot have an absolute measure of $\varphi_0$, hence we measure its fluctuation, $\Delta\varphi_0$. With the camera mounted on the stage, we collect a set of $100$ images of seven targets ($2\times 2$ checkerboard) acquired after the initialization and homing procedure of the stage, namely between two consecutive acquisitions we initialize the stage and send it to the home position. We detect the targets on the images with the subpixel routine in \cite{subPixel} that associates to each target the position of its central corner, and we measure the angular fluctuation of the home position within each pair of consecutive images as the displacement of the targets, normalized by the camera focal length $\Omega$.

To evaluate the natural fluctuations in the targets positions due to the detection routine, we perform a first test that we will use as a reference acquiring a set of $100$ images with the stage still in the home position. We compute the probability distribution function (PDF) of the angular fluctuations, highlighted in black in Fig.\ref{fig::home_repeatibility}. Then we perform the actual home repeatibility test acquiring a set of $100$ images with the stage in the home position, after performing the initialization and homing procedures, and we compute the PDF of this homing procedure fluctuations, highlighted in purple in Fig.\ref{fig::home_repeatibility}, which shows a zero median and values smaller than $6\cdot 10^{-5}$rad. 

The plot shows the high compatibility of the two PDFs, hence we conclude that the error on the home position is negligible, and smaller than the one guaranteed by the factory equal to $4\cdot 10^{-3}$rad.

\subsection{Angle interpolation}

In order to measure the error on the interpolation of the angles recorded by the stages, we perform the following test: we separate cameras and stages and we stuck on each stage a $13\times19$ checkerboard printed on a foam board. We put the stage in rotation, and we acquired images of the rotating checkerboard keeping the camera still, starting with the rotational stage in the \textit{home} position.
\begin{figure}[!h]
\centering
\includegraphics[width=1.0\linewidth]{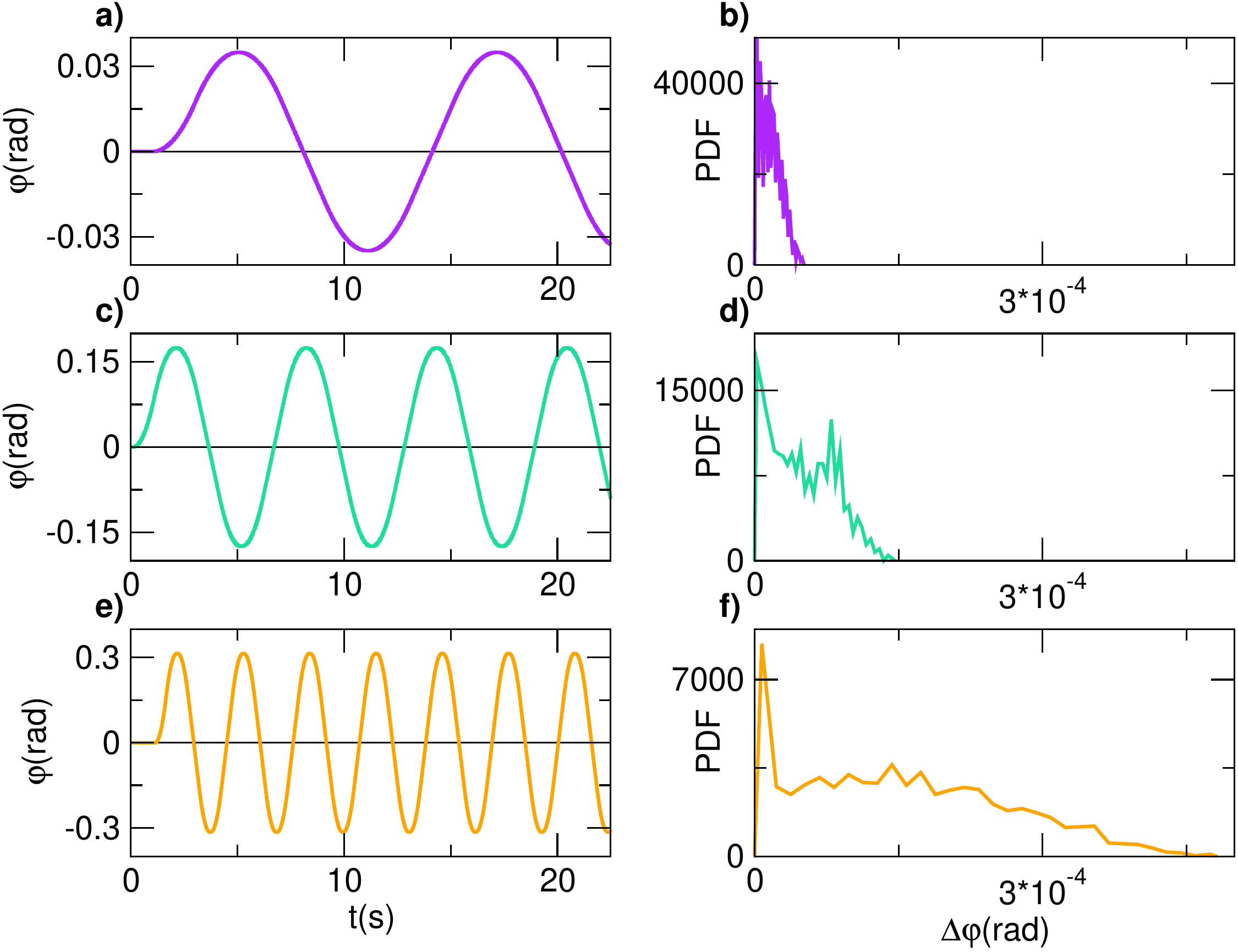}
\caption{\textbf{Angle accuracy.} \textbf{First column.} The angle gathered by the stage in the three different tests (slow, moderate and fast). \textbf{Second column.} PDFs (probability distribution function) of the error on the angle, $\Delta\varphi$, defined as the difference between the interpolation of the angle measured by the stage and the angle measured with Kabsch algorithm. The first row refers to the slow configuration, the second row refers to the moderate configuration and the third row to the fast configuration. As expected the error grows with the speed, due to the decrease of the interpolation accuracy with the speed, and in all the three cases the error is below $5\cdot10^{-4}$rad, being smaller than $5\cdot10^{-5}$rad for the slow configuration.}
\label{fig::stage_accuracy}
\end{figure}

We use the evolution in time of the position of the checkerboard corners to estimate the angle of rotation of the stage, in this way computing the rotation angle with a method that does not depend on the angular position gathered from the stage. 

For each image we detect the corners of the checkerboard with the subpixel routine in \cite{subPixel}. We use the first part of the acquisition, when the stage is still, to define a reference position for each corner, namely we associate to each corner the average of its coordinates over all the images with the stage in its home position. Then we compute the angle of rotation of the stage corresponding to a given camera frame, $t$, using Kabsch algorithm \cite{Kabsch}. More in detail, we associate to each frame $t$ the rotational matrix that minimizes the RMSD (root mean squared deviation), computed with Kabsch algorithm, between the positions of the corners detected at time $t$ and the reference positions. Finally, we compared the angle found with Kabsch algorithm and the angle that we would associate to the same camera frame interpolating the angular positions gathered by the stages. 

We carried out this test with the stages performing periodic rotation in three different configurations, corresponding to different choices of the parameters:

\noindent\textbf{1- slow.} $\varphi_{max}=2\degree$,  $v_{max}=1\degree/s$,  $a_{max}=0.5\degree/s^2$;

\noindent\textbf{2- moderate.} $\varphi_{max}=10\degree$,  $v_{max}=10\degree/s$, $a_{max}=10\degree/s^2$;

\noindent\textbf{3- fast.} $\varphi_{max}=18\degree$,  $v_{max}=36\degree/s$, $a_{max}=72\degree/s^2$. 

where $v_{max}$ and $a_{max}$ are the maximum speed and the maximum acceleration reached by the stages and $\varphi_{max}$ denote the amplitude of the periodic rotation, i.e. the stage performs a periodic rotation between $\varphi_{max}$ and $-\varphi_{max}$. 

The results of these tests are shown in Fig.\ref{fig::stage_accuracy} where in the first column we plot the angle gathered by the stage in the three different tests, and in the second column we show the PDF (probability distribution function) of the error on the angle, $\Delta\varphi$, defined as the difference between the interpolation of the angle measured by the stage and the angle measured via Kabsch algorithm. 

As expected we found that the error grows with the speed of the rotation, because of a decreasing accuracy in the interpolation, but in all cases we found an error smaller than $5\cdot10^{-4}$rad, being smaller than $5\cdot10^{-5}$rad for the slowest test, which can be considered negligible for all our practical purposes.

\section{System accuracy evaluation: $\mathbf{3D}$ tests}\label{sec::test3D}
The question at the very core of all $3D$ reconstruction systems is: how accurate is the system in reconstructing the position of an object $Q$ at a specific time $t$? Answering this question is not straightforward, especially if, as in our case, experiments are performed in the field where the system cannot be mounted and calibrated once and for all. We believe that a fair answer can only be given by checking reconstructed quantities against reality. 

This is what we actually do for our system in what we call $3D$ tests: with a laser range finder (Hilti Laser PD-E, accuracy $1$mm) we measure the distance between pair of targets in the common field of view of the cameras, we reconstruct the position of the targets in our world reference frame and from these positions we compute reconstructed target-to-target distances. Finally we compare reconstructed and measured distances and we compute the percentage error on the measured distances. 

We perform the $3D$ tests in two different fashion: \textit{i)} static $3D$ test. Cameras are set-up in their \textit{home} configuration and they do not move during the data acquisitions; \textit{ii)} dynamic $3D$ test. Cameras rotate during the data acquisition. 

\subsection{3D Test set-up}
We evaluate the $3D$ reconstruction accuracy of the system, checking that the requirements described in Section \ref{sec::requirements} are fulfilled, performing the tests described in detail in Section \ref{sec::CalibrationAccuracy}, Section \ref{sec::ImprovingOmega} and Section \ref{sec::alignmentTest}. In principle we should perform the tests exactly in the experimental configuration: camera baseline at $25$m, targets at a distance from the cameras in the range between $100$m and $150$m and pitch angles of both cameras set to $0.22$rad.

But due to logistic constraints we are forced to perform the tests in a slightly different configuration: \textit{i)} we set the camera baseline at about $10$m with targets at a distance from the cameras in the range between $20$m and $40$m; \textit{ii)} we do not manage to have targets in the common field of view of the cameras for a pitch value of $0.22$rad, but we can achieve the maximum pitch of $0.15$rad. We take care of these two logistic limitations in the design of the test and in the data analysis, see Section \ref{sec::SystemAccuracy}.

\subsection{Accuracy on the calibration of the internal and external parameters}\label{sec::CalibrationAccuracy}
To evaluate the accuracy of the calibration procedures we perform $3D$ tests in a special configuration where we can write the explicit coordinates of the reconstructed points. To this aim we set pitch and roll angles of both cameras equal to $0$, and we obtain the following explicit form of the $Z$ coordinate of a $3D$ point, see Appendix D:
\begin{equation}\label{eq::Z}
    Z(t) = \frac{\Omega d}{s(t)-(\alpha+\varphi(t))\Omega}
\end{equation}
where $d$ is the system baseline, i.e. the distance between the cameras that we measure with the laser range finder, $\Omega$ is the cameras focal length\footnote{For the sake of simplicity we are assuming the same value of the focal length for both cameras}, $s(t)=u_{\scaleto{L}{4pt}}(t)-u_{\scaleto{R}{4pt}}(t)$ is the disparity, $\alpha=\alpha_{\scaleto{R}{4pt}}-\alpha_{\scaleto{L}{4pt}}$ and $\varphi(t)=\varphi_{\scaleto{R}{4pt}}(t)-\varphi_{\scaleto{L}{4pt}}(t)$ are the mutual orientation of the cameras due to the system \textit{home} configuration and due to the rotation of the stages respectively.

From eq.(\ref{eq::Z}) we obtain the explicit expression of the relative error on $Z$, $\delta Z/Z$:
\begin{equation}\label{eq::dZ/Z}
    \frac{\delta Z(t)}{Z} = \frac{\delta d}{d} + \frac{\delta\Omega}{\Omega} + \frac{Z}{\Omega d}\left[\psi(t)\delta\Omega + \Omega\delta\psi(t)\right]
\end{equation}
where $\psi(t)=\alpha+\varphi(t)$ and $\delta d$, $\delta\Omega$ and $\delta\psi$ denote the error on $d$, $\Omega$ and $\psi$. Note that here we are not considering the contribute due to the error on $s(t)$, i.e. error in the position of the targets on the images, because it is not relevant in the $3D$ test set-up, see Appendix A.

Denoting the distance between two targets as $\Delta Z$ we obtain also that:
\begin{equation}\label{eq::dDZ/DZ}
    \frac{\delta(\Delta Z)}{\Delta Z} = \frac{\delta d}{d} + \frac{\delta\Omega}{\Omega} + 2\frac{\bar Z}{\Omega d}\left[\psi(t)\delta\Omega + \Omega\delta\psi(t)\right]
\end{equation}
where $\bar Z$ is the mean $Z$ coordinate of the two targets.

We do not measure the absolute positions of the targets but their mutual distances, $\Delta R$, hence in the $3D$ test we can only estimate the error on these distances, $\delta(\Delta R)$. In Appendix A we show that $\Delta R$ is proportional to $\Delta Z$, which means that $\delta(\Delta R)$ is proportional to $\delta(\Delta Z)$. Therefore we can write the explicit expression of the relative error on target-to-target distances by substituting $\delta(\Delta Z)/\Delta Z$ with $\delta(\Delta R)/\Delta R$ in eq.(\ref{eq::dDZ/DZ}), which gives:
\begin{equation}\label{eq::dDR/DR}
    \frac{\delta(\Delta R)}{\Delta R} = \frac{\delta d}{d} + \frac{\delta\Omega}{\Omega} + 2\frac{\bar Z}{\Omega d}\left[\psi(t)\delta\Omega + \Omega\delta\psi(t)\right]
\end{equation}
Eq.(\ref{eq::dDR/DR}) shows that $\delta(\Delta R)/\Delta R$ 
is made of a constant term, which depends on the error on $d$ and on $\Omega$, and of a linear term in $\bar Z$, which depends on the error on $\psi$ and $\Omega$. 
\begin{figure}[h]
    \centering
    \includegraphics[width=1.0\linewidth]{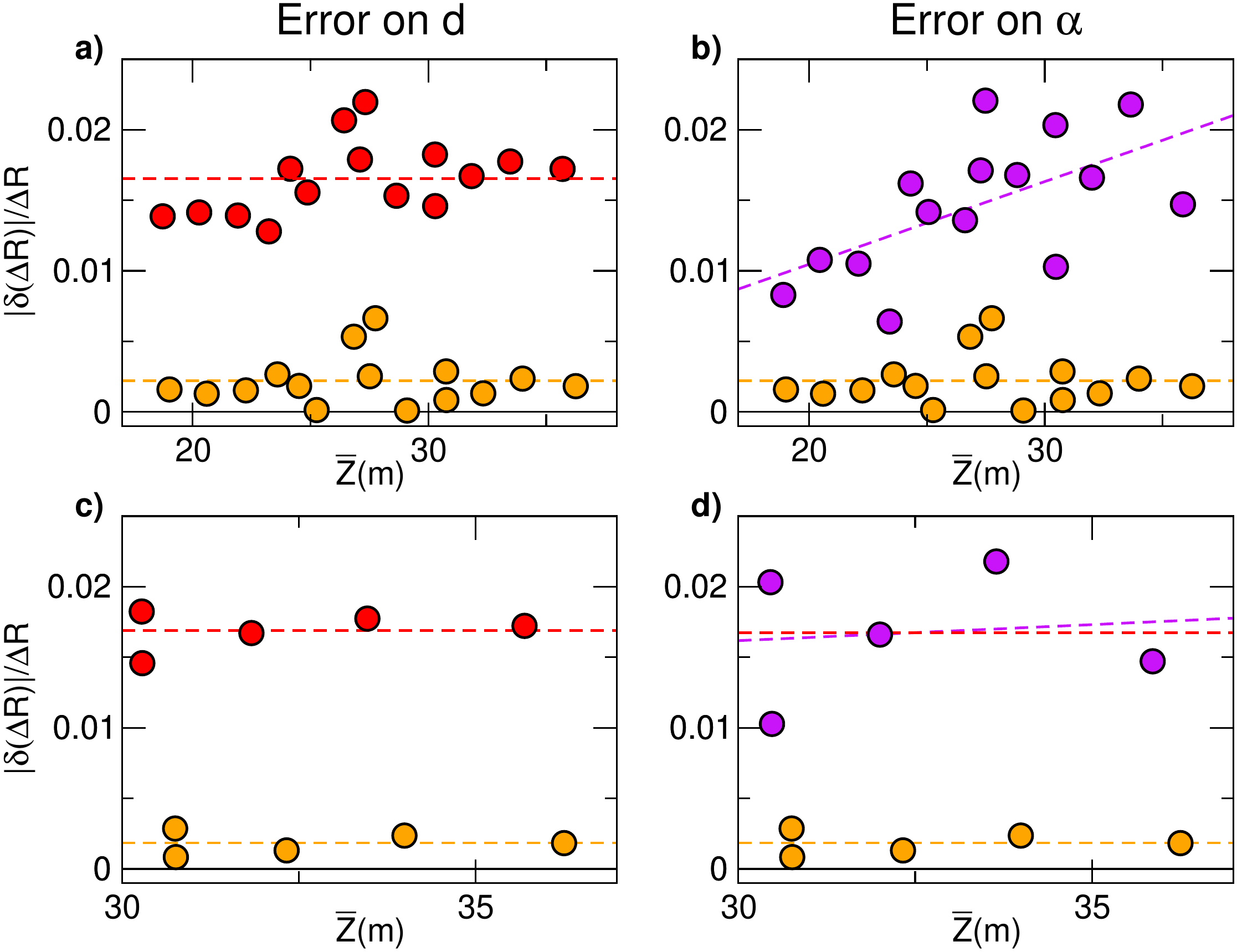}
    \caption{\textbf{Static} $\mathbf{3D}$ \textbf{test.} 
    The plots show $|\delta(\Delta R)|/\Delta R$ for each pair of targets as a function of their mean distance from the cameras, $\bar Z$. The orange circles represent the result of the $3D$ test obtained with the original calibration parameters. The orange dashed line is the mean value of $|\delta(\Delta R)|/\Delta R$. Red circles (left column) represent the results obtained by manually introducing an error of $0.1605$m in the baseline length ($\delta d/d=0.015$), while purple circles (right column) represent the result obtained by manually introducing an error of $0.003$rad in the angle $\alpha$. \textbf{a.} The error on $d$ produces an increment of the error, constant for all the targets. A constant fit of the data (red dashed line) gives an estimate of $\delta d/d$ equal to $0.016$, compatible with the experimental $\delta d/d=0.0015$. \textbf{b.} The error on $\alpha$ produces an increment of the error linear in $Z$. A linear fit of the data (purple dashed line) gives a slope equal to $0.00059 m^{-1}$, which corresponds to $\delta\alpha=0.0031$rad in perfect agreement with the experiment. \textbf{c.} When reducing the span of $Z$ the error due to $d$ is still well-estimated, with a constant fit that predicts an error on $\delta d/d$ equal to $0.016$ \textbf{d.} When reducing the span of $Z$ the error due to $\alpha$ cannot be detected and estimated properly. The constant fit (red dashed line) and the linear fit (purple dashed line) are both compatible with the data.} 
    \label{fig::Error_dAlpha}
\end{figure}

The idea now is to use the information of eq.(\ref{eq::dDR/DR}) to detect potential sources of error in the system.

From the trend of $\delta(\Delta R)/\Delta R$ in $\bar Z$ we can make a first discrimination between errors due to an incorrect measure of the baseline vs errors due to inaccuracies in $\Omega$ and $\psi$, as we show in Fig.\ref{fig::Error_dAlpha} where we present the effect on a static $3D$ test of an error on $d$ or of an error in $\alpha$. The difference between the two results is evident: an error on $d$ produces an increase of the errors constant with $Z$, while the error on $\alpha$ produces errors with a trend in $Z$. We stress here that it is possible to discriminate between the two situations only if the span in $Z$ of the targets is large enough, see Fig \ref{fig::Error_dAlpha} where in the bottom panels we show how the results would have looked like with a short span in $Z$.
\begin{figure*}[h]
    \centering
    \includegraphics[width=1.0\linewidth]{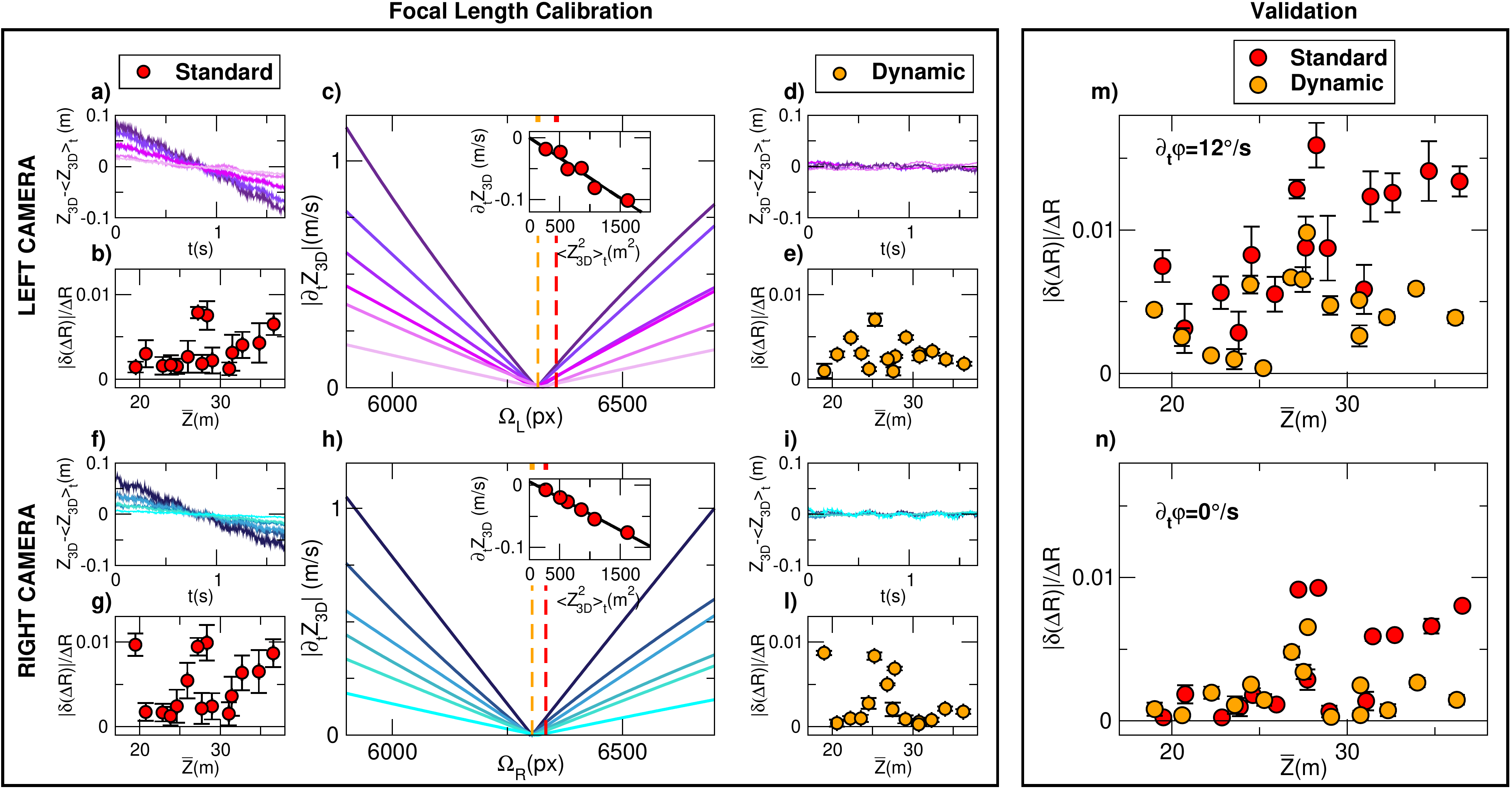}
    \caption{\textbf{Improving the focal length calibration.} In the left box data refer to the calibration improvement procedure, while on the right box to its validation. \textbf{Left box.} The top part refers to the calibration of the left camera and the bottom part to the calibration of the right camera. Data are collected with a dynamic $3D$ test with one camera per time in rotation at a constant speed $v=6\degree/s$. In the first column we show the results of the $3D$ test obtained with the standard $\Omega$, i.e. $\Omega$ calibrated with the standard method. In the right column we show the same quantities but obtained with the dynamic $\Omega$, i.e. $\Omega$ calibrated with the dynamic procedure. \textbf{a. d. f. and i.} Plots show the reconstructed $Z$, $Z_{3D}(t)$, for all the targets, each highlighted with a different color. $Z_{3D}(t)$ is normalized by its mean in time to have all the targets on the same range. \textbf{a. and f.} Standard $\Omega$: $Z(t)$ shows a linear trend in $t$. \textbf{d. and i.} Dynamic $\Omega$: $Z(t)$ does not show any trend in $t$. \textbf{b. e. g. and l.} Plots show the mean in time of $\delta(\Delta R)/\Delta R$ for each pair of targets as a function of the pairs mean distance from the cameras, $\bar Z$. Error bars are computed as standard deviation. \textbf{b. and g.} Standard $\Omega$: large error bars reflect the high variability of the targets $Z(t)$ due to their linear trend in $t$. \textbf{d. and l.} Dynamic $\Omega$: error bars are in most of the cases smaller than the symbols and they reflect the absence of the trend  in time of the targets $Z(t)$. \textbf{c. and h.} The plots show the absolute value of the slope of the reconstructed $Z$, $|\partial_t Z_{3D}(t)|$, as a function of $\Omega$. At a fixed value of $\Omega$ the slope increases with the target distance from the camera, which is embedded in the color code, going from light purple and light blue for the closest target to dark purple and dark blue for the furthest. All the targets present a well-defined minimum of the slope for the same value of $\Omega$, highlighted with an orange dashed line, which corresponds to the dynamic $\Omega$, while the standard $\Omega$ is highlighted with the red dashed line. In the inset we show the linear trend of $\partial_t Z_{3D}$ with the average of $Z_{3D}$ in time, $<Z^2_{3D}>_t$ for the standard $\Omega$. \textbf{Right box.} We validate the dynamic calibration comparing the absolute value of the relative error in the target-to-target distances using the focal length obtained with the standard (red circles) and dynamic (orange circles) calibration. \textbf{m.} We tested the dynamic calibration with a dynamic $3D$ test rotating both cameras simultaneously at a speed of $6\degree/s$ in the two opposite directions. The plot shows that with the dynamic calibration we obtain smaller relative errors and much smaller error bars than with the standard calibration. Moreover we see that the trend in $Z$ that is quite evident for the standard calibration, becomes negligible with the dynamic calibration. \textbf{n.} We validate the dynamic calibration on a $3D$ test reproducing our experimental procedure, with both cameras rotating simultaneously and in the same direction. Here we do not appreciate a decrease of the error bars, because the effect of $\delta\Omega$ is negligible due to the effective speed $v=0\degree/s$, but we still see that the overall errors get smaller.
    }
    \label{fig::OmegaCalibration}
\end{figure*}
To further discriminate between an error in $\Omega$ and an error in $\psi(t)$ we need dynamic information. To this aim we derive $Z$ with respect to time and we obtain the following expression:
\begin{equation}\label{eq::dtdZ}
    \partial_t(\delta Z) = \frac{Z^2}{\Omega d}\left[\partial_t\varphi(t)\cdot\delta\Omega\right]
\end{equation}
which tells us that the evolution in time of the error on $Z$ is quadratic in $Z$ with a coefficient that depends on the rotational speed, $\partial_t\varphi$, and on the error on the focal length, $\delta\Omega$.

In Section \ref{sec::InternalCalibration} we mentioned that we need a two-steps procedure for the calibration of the internal parameters, because of the low accuracy on the estimation of $\Omega$ with the standard calibration approach. We will use this last equation to show how to detect and how to quantify the error $\delta\Omega$. Once we corrected the error on $\Omega$ we can go back to eq.(\ref{eq::dDR/DR}) and check for a potential error on the cameras orientation, with the tests described in Section \ref{sec::CalibrationAccuracy}.

\subsubsection{Improving the focal length calibration}\label{sec::ImprovingOmega}

\begin{figure*}[h]
    \centering
    \includegraphics[width=1.0\linewidth]{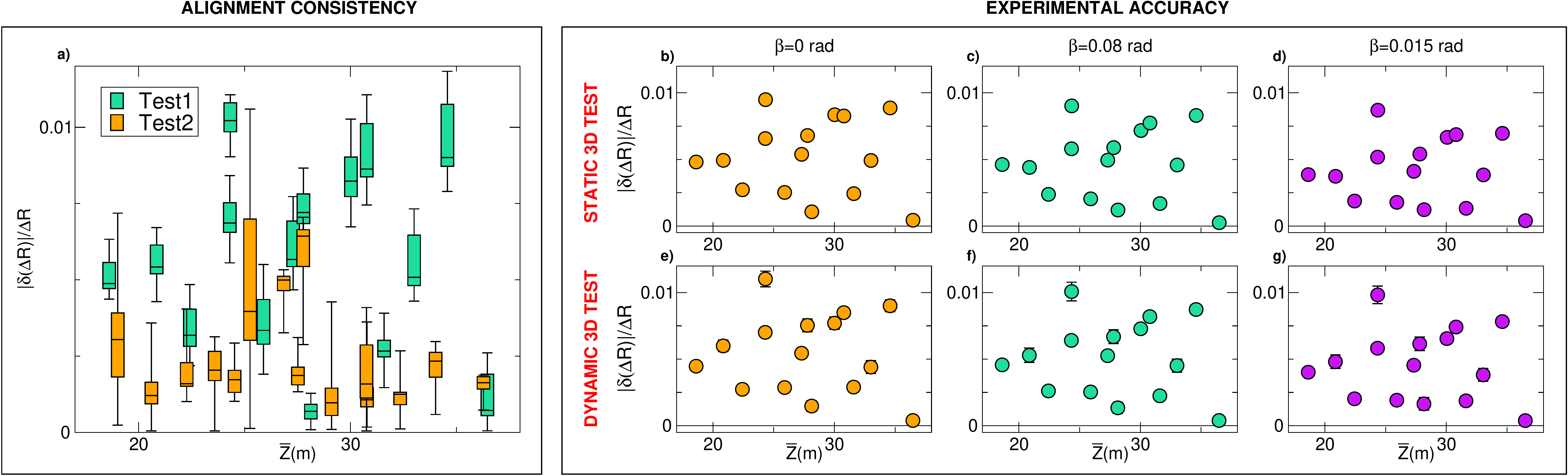}
    \caption{\textbf{System accuracy.} \textbf{Left box.} The orange and the green boxplots represent the relative error in the $3D$ reconstruction as a function of $\bar Z$ for two different sets of static $3D$ tests. Data from the two sets are collected mounting and unmounting the entire system, which justify differences in the $\bar Z$ values for the two tests. Data within the same set are collected by repeating the alignment procedure with the fishing line ($8$ times for the orange set and $10$ times for the purple one). The line inside the box correspond to the median of the relative error in the $3D$ reconstruction for a single target-to-target distance, the two edges of the box correspond to the first and the third quartiles, the two whiskers correspond to the minimum and the maximum value.  The plot show no trend in $\bar Z$, hence showing a not appreciable error on the angle $\alpha$. The data show variability within the same test (quite large error bars), due to the alignment, and also a variability within the two different tests, due to the set-up procedure, but this does not affect the accuracy and the consistency of the $3D$ reconstruction that gives always relative errors smaller than $0.012$.\textbf{Right box.} Data presented in the first and the second row are collected performing respectively static and dynamic $3D$ tests for different values of $\beta$. The dynamic tests are performed in the field configuration with the cameras rotating simultaneously at the same speed and in the same direction. The plots show $|\delta(\Delta R)|/\Delta R$ for each pair of targets as a function of their mean distance from the cameras, $\bar Z$. Static tests are performed shooting one single image, hence we do not have error bars. For the dynamic tests instead, we plot the $|\delta(\Delta R)|/\Delta R$ averaged in time and error bars, which are most of the times smaller than the symbols, represent standard deviation. We do not see any trend of the error with $\beta$ nor in the static tests nor in the dynamic tests. The comparison between static and dynamic tests at a fixed value of $\beta$ show relative errors of the same order and always smaller than $0.01$.} 
    \label{fig::SystemAccuracy}
\end{figure*}

We check the accuracy of the standard calibration of $\Omega$ with the following $3D$ test: we put in rotation one camera per time at a constant rotational speed ($v=6\degree/s$). We check $\Omega$ of the left camera rotating only the left camera in the clockwise direction:
\begin{equation}\label{eq::dtphi_L}
    \partial_t\varphi_{\scaleto{L}{4pt}}(t)=v\mbox{ and } \partial_t\varphi_{\scaleto{R}{4pt}}(t)=0
\end{equation}
while we check  $\Omega$ of the right camera rotating only the right camera in the counterclockwise direction:
\begin{equation}\label{eq::dtphi_R}
    \partial_t\varphi_{\scaleto{L}{4pt}}(t)=0\mbox{ and } \partial_t\varphi_{\scaleto{R}{4pt}}(t)=-v
\end{equation}
Therefore in both tests $\partial_t\varphi(t)=\partial_t\varphi_{\scaleto{R}{4pt}}(t)-\partial_t\varphi_{\scaleto{L}{4pt}}(t)=-v$,
and eq.(\ref{eq::dtdZ}) reads:
\begin{equation}\label{eq::dtdZ_v}
\partial_t(\delta Z(t))=-v\frac{Z^2}{\Omega d}\delta\Omega.
\end{equation}
Note that $\delta Z$ is the reconstruction error, hence $\delta Z(t)=Z_{3D}(t)-Z$ where $Z_{3D}$ is the reconstructed $Z$. This implies that $\partial_t(\delta Z(t))=\partial_t Z_{3D}(t)-\partial_t Z$ but the targets are still, hence their position is constant in time and $\partial_t Z=0$. Eq.(\ref{eq::dtdZ_v}) can then be written as:
\begin{equation}\label{eq::dtZ3D}
\partial_t(Z_{3D}(t))=-v\frac{Z^2}{\Omega d}\delta\Omega
\end{equation}
which tells us that the derivative of $Z_{3D}$ with respect to time is constant for each target and it linearly depends on the speed of rotation and on the error in $\Omega$. We checked the evolution in time of $Z_{3D}(t)$ and we found the linear trend shown Fig.\ref{fig::OmegaCalibration}a and Fig.\ref{fig::OmegaCalibration}f, which is also the reason for the large error bars of $\delta(\Delta R)/\Delta R$ in Fig.\ref{fig::OmegaCalibration}b and Fig.\ref{fig::OmegaCalibration}g.

Eq.(\ref{eq::dtZ3D}) tells us more, because it shows that $\partial_t Z_{3D}(t)$ is quadratic in $Z$, which means that targets at different distances from the cameras will have a linear trend in time with different slopes: the further apart the target the higher the slope. With a linear fit of $Z_{3D}(t)$ we computed $\partial_tZ_{3D}(t)$ for each target and we plot these quantities versus $<Z_{3D}^2>_t$\footnote{$<Z_{3D}^2>_t$ is the average in time of $Z_{3D}^2(t)$ and it is the most accurate estimate of $Z$ that we can give, since we do not measure the absolute position of the targets but targets mutual distances.}, see insets in Fig.\ref{fig::OmegaCalibration}c and Fig.\ref{fig::OmegaCalibration}h. From these last plots we estimated $\delta\Omega$ with a linear fit, and we found an error of $41.61$px for the left camera ($\Omega_L=6356.41$px with the standard calibration and $\Omega_L=6314.8$px with the dynamic calibration) and an error of $33.52$px for the right camera ($\Omega_R=6333.81$px with the standard calibration and $\Omega_R=6300.29$px with the dynamic calibration). 

But we can estimate $\delta\Omega$ more precisely with a different strategy: we run again the analysis of the $3D$ test moving the value of $\Omega$ in the interval $[5900px, 6700px]$ and, for each value of $\Omega$, we compute $|\partial_t Z_{3D}(t)|$ of each target. We found that all the targets have a well-defined minimum of $|\partial_t Z_{3D}(t)|$ that occurs at the same value of $\Omega$, see Fig.\ref{fig::OmegaCalibration}c and Fig.\ref{fig::OmegaCalibration}h. We choose then $\Omega$ corresponding to this minimum as our new calibrated focal length, i.e. the dynamic $\Omega$ highlighted with a dashed orange line in Fig.\ref{fig::OmegaCalibration}c and Fig.\ref{fig::OmegaCalibration}h. With this procedure we found an error on $\Omega$ for the left camera equal to $40$px and for the right camera equal to $30$px, compatible with the estimate obtained from the linear fit of $\partial_tZ_{3D}$ vs $Z_{3D}^2$. We checked that using this dynamic $\Omega$, $Z_{3D}$ does not show anymore a trend in $t$  and we also found a reduction of the error bars for $\delta(\Delta R)/\Delta R$, as shown in Fig.\ref{fig::OmegaCalibration}d, Fig.\ref{fig::OmegaCalibration}e, Fig.\ref{fig::OmegaCalibration}i and Fig.\ref{fig::OmegaCalibration}l.

We validated the dynamic calibration performing other two $3D$ tests in different conditions. We perform a first test rotating both the cameras simultaneously at a constant speed of $6\degree/s$ but in opposite directions, in this way amplifying a potential error on $\Omega$: we rotate the left camera in the clockwise direction, $\partial_t\varphi_{\scaleto{L}{4pt}}(t)=-v$ and the right camera in the counterclockwise direction, $\partial_t\varphi_{\scaleto{R}{4pt}}(t)=v$, hence the effective rotational speed $\partial_t\varphi$ is equal to $12\degree/s$.
We also performed a second test to simulate the experimental set-up, thus rotating the cameras in the same direction at the same speed, $\partial_t\varphi_{\scaleto{L}{4pt}}(t)=\partial_t\varphi_{\scaleto{R}{4pt}}(t)=v$ with an effective rotational speed $\partial_t\varphi(t)=0\degree/s$.

The results of these two tests are shown in Fig.\ref{fig::OmegaCalibration}m and Fig.\ref{fig::OmegaCalibration}n, where the red circles refer to the standard calibration and the orange circles to the dynamic calibration. As expected the effect of the standard $\Omega$ is more evident in the test at $\partial_t\varphi(t)=12\degree/s$ where we see large error bars of $|\delta(\Delta R)|/\Delta R$ and also a trend with $\bar Z$, while for $v=0\degree/s$ error bars are quite small. In both tests the dynamic $\Omega$ reduces $|\delta(\Delta R)|/\Delta R$ and it makes the error bars for the test at $v=12\degree/s$ comparable with the ones of the test at $\partial_t\varphi(t)=0\degree/s$. These two factors, lower $|\delta(\Delta R)|/\Delta R$ and smaller error bars, confirm that the dynamic $\Omega$ is more correct than the one obtained with the standard calibration.

From these tests we learn that for an accurate calibration of the internal parameters, we need first to perform the standard calibration procedure described in Section \ref{sec::InternalCalibration} and then we need to perform two dynamic $3D$ tests, each with only one camera per time in rotation at a constant speed. From the linear fit of $\partial_t Z_{3D}(t)$ versus $<Z_{3D}^2>_t$ we estimate the error on the focal length of the two cameras, which we use to correct the results obtained with the standard calibration approach. With this two-steps calibration procedure, we fulfill the requirement on the time independence of the reconstruction error at the relatively low cost of performing two dynamic $3D$ tests, namely few hours of work.

\subsubsection{Set-up and alignment consistency}\label{sec::alignmentTest}
Field experiments are often performed in locations where the apparatus cannot be mounted once and for all as it happens for our experiment, which is carried out on the roof of a building where we are forced to mount and unmount the entire system on a daily basis. It is then important to design an easy-to-mount system and a consistent calibration procedure. We tested CoMo to evaluate our consistency in the mounting procedure and in the alignment of the cameras with the fishing line, described in Fig.\ref{fig::experimentalSetup}. 

To this aim we performed two sets of static $3D$ test mounting and unmonting the entire system between the two. In each set we repeat several times the alignment procedure taking at every alignment a static picture of the targets. We then reconstructed the position of the targets, we computed target-to-targets distances and $\delta(\Delta R)/\Delta R$, Finally we evaluate the variability of the reconstruction error within each set of data and between the two  sets.

The results of this test are shown in Fig.\ref{fig::SystemAccuracy}a, where we show the relative error, $\partial(\Delta R)/\Delta r$, of each pair of targets as a function of $\bar Z$. The plot shows variability within the same test, which is due to the alignment procedure, and variability within different tests, but with relative errors always below 0.012. The absence in both tests of a trend in $\bar Z$ shows that inaccuracies in the calibration of $\alpha$ are negligible and that the alignment technique is consistent, while the upper limit of $0.0012$ of the reconstruction error shows the consistency of our mounting procedure.

\subsection{ $\mathbf{3D}$ reconstruction accuracy in field set-up}\label{sec::SystemAccuracy}
We evaluate the $3D$ reconstruction accuracy of the system performing again $3D$ tests, but this time with a set-up as similar as possible to the experimental one.
In principle we should perform this $3D$ test exactly in the experimental configuration: camera baseline at $25$m, targets at a distance from the cameras in the range between $100$m and $150$m and pitch angles of both cameras set to $0.22$rad.

But due to logistic constraints we are forced to perform the tests in a slightly different configuration: \textit{i)} we set the camera baseline at about $10$m with targets at a distance from the cameras in the range between $20$m and $40$m; \textit{ii)} we do not manage to have targets in the common field of view of the cameras for a pitch value of $0.22$rad, but we can achieve the maximum pitch of $0.15$rad. 

We take care of these two logistic limitations in the design of the test and in the data analysis. In particular: \textit{i)} in the $3D$ test the ratio $Z/d$ is between $2$ and $4$, while this ratio in the field is between $4$ and $6$. The factor $Z/d$ is relevant when we find a trend with $\bar Z$ in $\delta(\Delta R)/\Delta R$ in which case, to estimate the experimental error, we have to renormalize the $3D$ reconstruction error found in the tests by a factor $2$; \textit{ii)} we perform three series of tests at different pitch angles: $\beta=0$rad, $\beta=0.08$rad and $\beta=0.15$rad to detect a potential trend of the error with $\beta$ and in this case to predict the range of the reconstruction error in the field conditions, i.e. $\beta=0.22$rad.

We perform the $3D$ test in the following way: for each of the three pitch values we perform first a static $3D$ test in the \textit{home} configuration and then we put both cameras in rotation as in the field, namely both cameras rotate in the same direction and with the same rotational speed. We perform the test rotating the cameras at a constant speed of $6\degree/s$ $(0.1rad/s)$, which is the maximum speed we use in the field.

The results are shown in the right box of Fig.\ref{fig::SystemAccuracy},  where in the first row we plot the relative error for the three static tests and in the second row the results of the dynamic $3D$ tests. In both cases we have excellent results with relative errors smaller than $0.01$, without any trend in $Z$ and we did not find any trend of the error with $\beta$. We do not have here to renormalize the relative error to take care of the different value of $Z/d$ in the test and in the field because we do not see any trend of the relative error in $Z$, nor we need to make any prediction of the error at $\beta=0.22$rad because there are not appreciable differences of the errors for different $\beta$. 

The results of the static $3D$ test at $\beta=0$ essentially reflects the accuracy of the cameras alignment procedure. The comparison between static tests at different values of $\beta$ shows that the introduction of a non-zero pitch angle produces a negligible error, because with different values of $\beta$ we obtain errors of the same order. With a similar argument, the comparison between static and dynamic tests shows that the introduction of the rotation due to the stages does not add affect the accuracy of the external parameters calibration. Therefore the $3D$ tests shows that the dominant source of error on the external parameters calibration is the alignment technique and in particular on the measurement of the cameras yaw angles. From the results of the $3D$ tests we estimated this angular error to be smaller than $0.001$rad, hence confirming the high-precision of our alignment procedure.

\section{Conclusions}
We presented a novel co-moving camera stereo system, CoMo, developed in the context of $3D$ tracking of large groups of targets moving in a wide and non-confined space. To overcome the limitation of standard static set-up, where the size of the field of view is defined by the fixed position of the cameras and in most of the cases narrowed to achieve a sufficient resolution of the system, we designed CoMo to follow the motion of the targets with a controlled and synchronized rotation of the cameras driven by rotational stages (one for each camera).

The $3D$ reconstruction for a dynamic and wide field system is rather demanding because the external parameters of the system have to be calibrated frame-by-frame and they cannot be calibrated with standard methods, which are not accurately enough on wide field data. We propose a novel technique for the calibration of the external parameters that separates their static component, corresponding to the system in the \textit{home} configuration (rotational stages at the $0\degree$ position), from their dynamic component, corresponding to the rotation due to the stages. We calibrate the static component of the external parameters by measuring the position and the three angles of yaw, pitch and roll of the cameras in a common reference frame, and we combine this information with the frame-by-frame rotation gathered from the stages.

We validated this calibration approach performing what we call \textit{$3D$ tests}: we set-up the system, we acquire images of a set of still targets, and we accurately measure with a laser distometer the distance between each pair of target. From the collected images we reconstruct the position of the targets and we compute their mutual distances that we compare with the measure ones. The results of the $3D$ tests show the consistency of the calibration method for the external parameters and the high accuracy of the system ($3D$ reconstruction error below $1\%$). 

$3D$ tests represent a fair and objective method to evaluate the accuracy of a $3D$ system but the very relevance of the $3D$ tests is in the designing phase of a $3D$ system because, as we showed in the manuscript, $3D$ tests are a powerful tool to detect potential sources of errors also providing a well-defined procedure to discriminate errors due to an incorrect measurement of the cameras position vs errors due to an incorrect measurement of the cameras orientation. Finally, $3D$ tests are at the basis of the new method that we proposed to improve the standard calibration of the focal length, which we could found to be inaccurate by performing dynamic $3D$ tests and noting an unexpected trend of the reconstructed position with time. 

We carried a first experimental campaign using CoMo to collect data on starling flocks, which are an emblematic example of targets moving in large groups in a non-confined space. To this aim we set-up the apparatus on the roof of Palazzo Massimo alle Terme, where we are forced to mount and unmount the system every day. With this first experimental campaign we proved that  the system is easy to mount and easy to calibrate and confirmed that the design of CoMo considerably expand the time-length of the acquired data.




\section*{Acknowledgments}
We thank Zachary Stamler for the fruitful and stimulating discussion about the calibration of the external parameters of the system. This work was supported by ERC grant RG.BIO (Grant No. 785932).

\bibliographystyle{IEEEtran}
\bibliography{IEEEabrv,biblio}



\end{document}